\documentclass[review]{elsarticle}

\usepackage{lineno,hyperref}

\usepackage{amssymb,amsmath,amsfonts}
\usepackage{multirow, array, graphicx, epsfig, lineno, subfigure}
\usepackage{xcolor}
\modulolinenumbers[5]

\journal{Information Fusion}

\begin{document}

\begin{frontmatter}

\title{Improving Automated Latent Fingerprint Identification\\ using
  Extended Minutia Types\footnote{\textbf{© 2018. This manuscript version is made available under the CC-BY-NC-ND 4.0 license.
                            http://creativecommons.org/licenses/by-nc-nd/4.0/ \\
     The final version is available online at: https://doi.org/10.1016/j.inffus.2018.10.001}}}

\author[address1]{Ram P. Krish}\ead{ram.krish@dcu.ie}
\author[address2]{Julian Fierrez\corref{cor}}
\cortext[cor]{Corresponding author}
\ead{julian.fierrez@uam.es}
\author[address2]{Daniel Ramos}\ead{daniel.ramos@uam.es}
\author[address3]{Fernando Alonso-Fernandez}\ead{feralo@hh.se}
\author[address3]{Josef Bigun}\ead{josef.bigun@hh.se}

\address[address1]{School of Electronic
  Engineering, Dublin City University, Ireland\\}

\address[address2]{School of Engineering, Universidad Autonoma de
  Madrid, Spain}

\address[address3]{School of Information Technology, Halmstad
  University, Sweden}

%-------------------------------------------------------------------------
\begin{abstract}
%\scriptsize
\emph{
  Latent fingerprints are usually processed with
  Automated Fingerprint Identification Systems (AFIS) by law
  enforcement agencies to narrow down possible suspects from a
  criminal database.
  AFIS do not commonly use all discriminatory features available in fingerprints
  but typically use only some types of features automatically extracted by a
  feature extraction algorithm.
  In this work, we explore ways to improve rank identification
  accuracies of AFIS when only a partial latent fingerprint is
  available.
  Towards solving this challenge, we propose a method that exploits
  extended fingerprint features (unusual/rare minutiae) not commonly considered
  in AFIS. This new method can be combined with any existing
  minutiae-based matcher.
  We first compute a similarity score based on least squares between
  latent and tenprint minutiae points, with rare minutiae features as
  reference points.
  Then the similarity score of the reference minutiae-based matcher at
  hand is modified
  based on a fitting error from the least square similarity stage.
  We use a realistic forensic fingerprint casework database in our
  experiments which contains
  rare minutiae features obtained from Guardia
  Civil, the Spanish law enforcement agency.
  Experiments are conducted using three minutiae-based matchers as a
  reference, namely:
  NIST-Bozorth3, VeriFinger-SDK and MCC-SDK. We report significant improvements in
  the rank identification accuracies when these minutiae matchers are
  augmented with our proposed algorithm based on rare minutiae features.\\
}

\end{abstract}
%-------------------------------------------------------------------------
%-------------------------------------------------------------------------

\end{frontmatter}

%\linenumbers

%-------------------------------------------------------------------------

\section{Introduction}\label{sect:introduction}

Fingerprints left at a crime scene, referred to as latent prints, are
the most common type of forensic science evidence and have been used
in criminal investigations for more than 100 years, but comparing latent
fingerprints is not an easy task. This is mainly attributed to the
poor quality of the latent fingerprints obtained from the crime
scenes. When a latent fingerprint is found,
the criminal investigators first search for the suspect
in their criminal database using an
Automated Fingerprint Identification System (AFIS) to
narrow down their manual work. If there is a match, then the
individual is linked to the crime under investigation.
Individualization (\emph{identification or match}) is the decision
yielded by a forensic examiner about the latent fingerprint
belonging to a particular individual. This is the outcome of the
Analysis, Comparison, Evaluation and Verification (ACE-V)~\citep{acev}
methodology currently followed in friction ridge examination.

\begin{figure}[t]
  \centering
  \subfigure[Good]
            {\includegraphics[width=3.9cm]{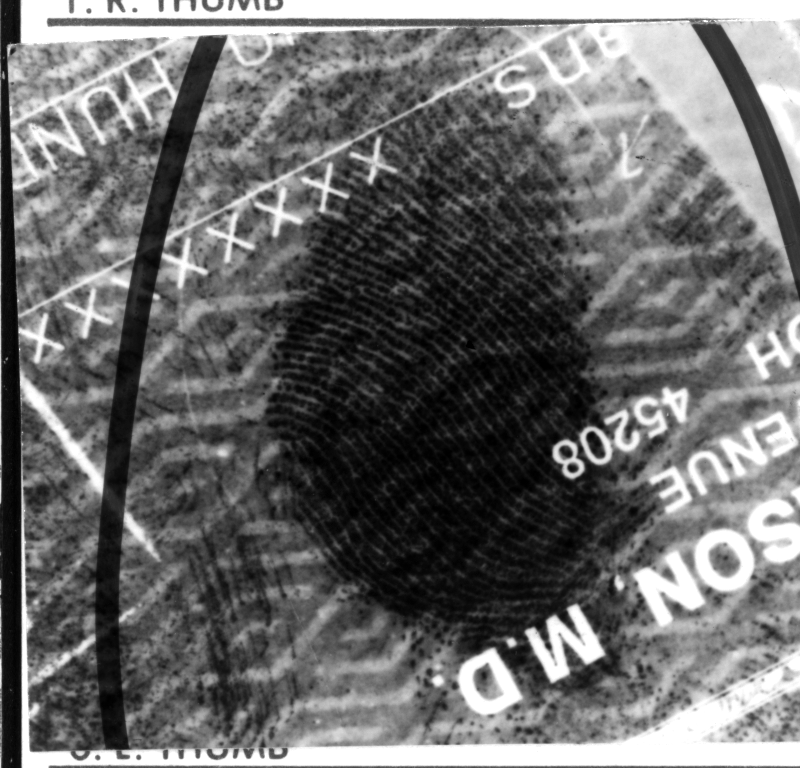}}
 \subfigure[Bad]
            {\includegraphics[width=3.9cm]{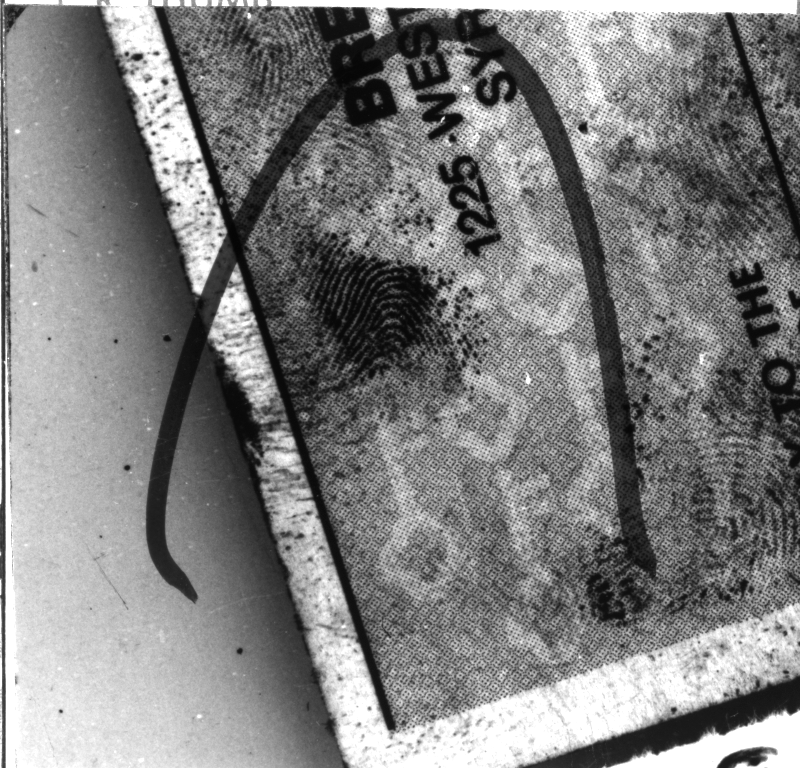}}
 \subfigure[Ugly]
            {\includegraphics[width=3.9cm]{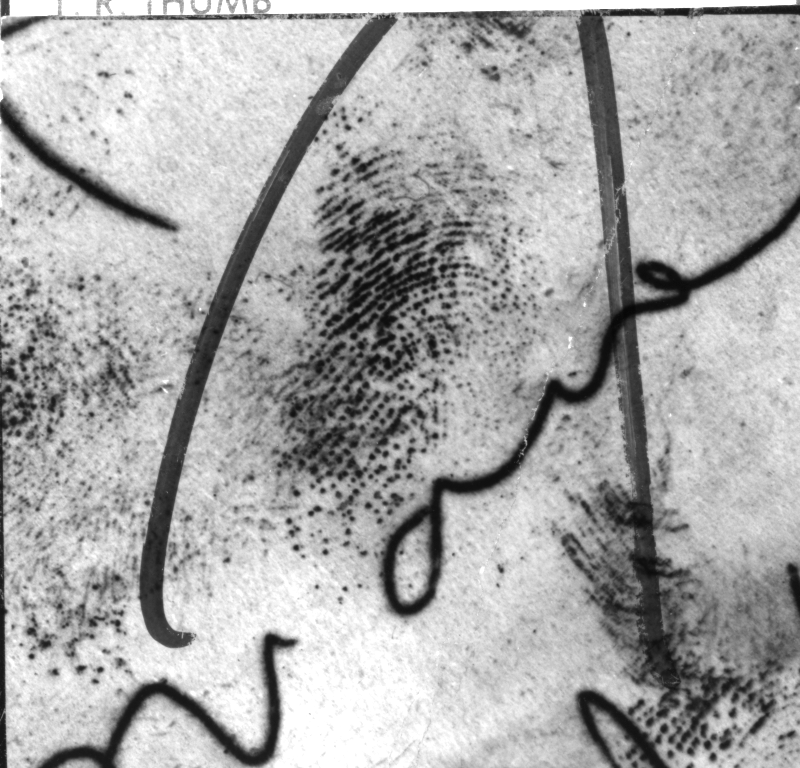}}
  \caption{Subjective quality classification of latent fingerprint
    images in NIST Special Database 27 (NIST-SD27).}
  \label{fig:gbuImg}

\end{figure}

In order to improve the matching
efficiency, the concept of ``Lights-Out System'' was
introduced for latent matching~\citep{lightsout}.
A Lights-Out System is a fully automatic
identification process with no human intervention.
Here, the system should automatically extract
the features from the latent fingerprint and match it against the
tenprints (exemplars) stored in the AFIS database to obtain a set of possible
suspects with high degree of confidence. In
general, latent fingerprints are partial in nature and are of varying
quality (see Figure~\ref{fig:gbuImg}), mostly
distorted, smudgy, blurred, etc.
These factors lead to high number of
unreliable extracted features in fully automatic mode, and make it
difficult for AFIS to perform well.
AFIS do not commonly use all the discriminatory
features that could be derived from a fingerprint, mainly due to the
limitations of automatic and reliable extraction of all types of
discriminatory features. The accurate performance of feature
extraction and matching algorithms of AFIS in forensic scenario is of
great importance to avoid erroneous individualization.

Current practice in latent AFIS technology involves marking the latent
fingerprint features manually by forensic examiners and then using both the latent
fingerprint image and the manually marked features to search in the
AFIS for a list of suspects. Sankaran et al.~\citep{anush01} reviewed exisiting automated
latent fingerprint matching algorithms and their limitations.
To avoid the burden of manual
marking and with the hope of fully automating the latent AFIS in Lights-Out
mode, the US National Institute of Standards \& Technology
(NIST) conducted a public evaluation of commercial AFIS
performance in Lights-Out mode. This was a multi-phase open project
called Evaluation of Latent Fingerprint Technologies
(ELFT)~\citep{nistELFT}. The results of various phases of ELFT are
summarized in Table~\ref{tab:elft4}. The reported accuracies from
Phase-I and Phase-II of ELFT cannot be directly compared as the database and
the quality of the latents were different.
As one of the results of the ELFT initiative~\citep{elft4}, it is
shown that in some practical conditions manual intervention may be
needed, as fully automated procedures are not yet robust enough,
especially when challenging latent fingerprints are considered.
The procedures of marking the minutiae, determining the subjective
quality of latents, etc. may need to be carried out manually.

\begin{table}[tbh]
\footnotesize
\begin{center}
\begin{tabular}{|p{3.2cm}|p{4cm}|p{1.5cm}|}
\hline
\vspace{0.005in}\bf{\hspace{0.3in}Phase of ELFT}
& \vspace{0.005in}\bf{\hspace{0.5in}Database size} &
\bf{Rank-1\newline accuracy}\\
\hline\hline
\vspace{0.1in}Phase-I \cite{elft1} & \vspace{0.005in} 100 latents
compared against 10,000 rolled prints & \vspace{0.005in}$80.0\%$ \\
\hline
\vspace{0.1in}Phase-II, Evaluation-1 \cite{elft2} & \vspace{0.005in} 835 latents
compared against 100,000 rolled prints & \vspace{0.005in}$97.2\%$ \\
\hline
\vspace{0.1in}Phase-II, Evaluation-2 \cite{elft3} & \vspace{0.005in} 1,114 latents
compared against 100,000 rolled prints & \vspace{0.005in}$63.4\%$ \\
\hline
\end{tabular}
\end{center}
\caption{Summary of NIST Evaluation of Latent Fingerprint Technologies
(ELFT) results.}
\label{tab:elft4}
\end{table}

AFIS commonly use only a limited number of features
automatically extracted from the fingerprints using a feature
extraction algorithm. On the other hand, forensic examiners use a richer
set of features during their
manual comparisons. This could be
a possible reason why manual comparisons outperform AFIS
comparisons~\citep{jainEFS}. Any features that are not commonly used
by commercial AFIS are generally termed as Extended Feature Sets
(EFS)~\citep{elft2}. The use of EFS by forensic examiners in manual
comparisons is much debated, mainly due to non-repeatability
by another examiner to validate the previous decision.

Two major problems in friction ridge analysis that raised the
attention of the forensic community about 10 years ago were identified
as follows in~\citep{stdEFS}:
\begin{enumerate}
  \item Latent AFIS searches are commonly limited by an over simplified feature set.
  \item During the latent examiner comparison, there are no standard
    format to document the features used in comparison decision. This
    leads to problems with future reference or interchange with
    other forensic examiners.
\end{enumerate}

The SWGFAST (Scientific Working Group on Friction Ridge Analysis, Study,
and Technology) drafted a memo to NIST noting that forensic examiners
use features that are not currently addressed in fingerprint
standards. The ANSI/NIST Standard Workshop II charted the Committee
to Define an Extended Fingerprint Feature Set (CDEFFS). The CDEFFS
included $45$ members from various federal agencies, the forensic community,
AFIS vendors, and academia~\citep{stdEFS}. The purpose of CDEFFS was to
define a standard to completely represent the distinctive information
in the fingerprint which are quantifiable, repeatable and develop a clear
method of characterizing information for: 1) AFIS searches initiated
by forensic examiner, and 2) forensic examiner markup and exchange
of latent fingerprints.

\begin{figure}[t]
\centering
\includegraphics[scale=0.3]{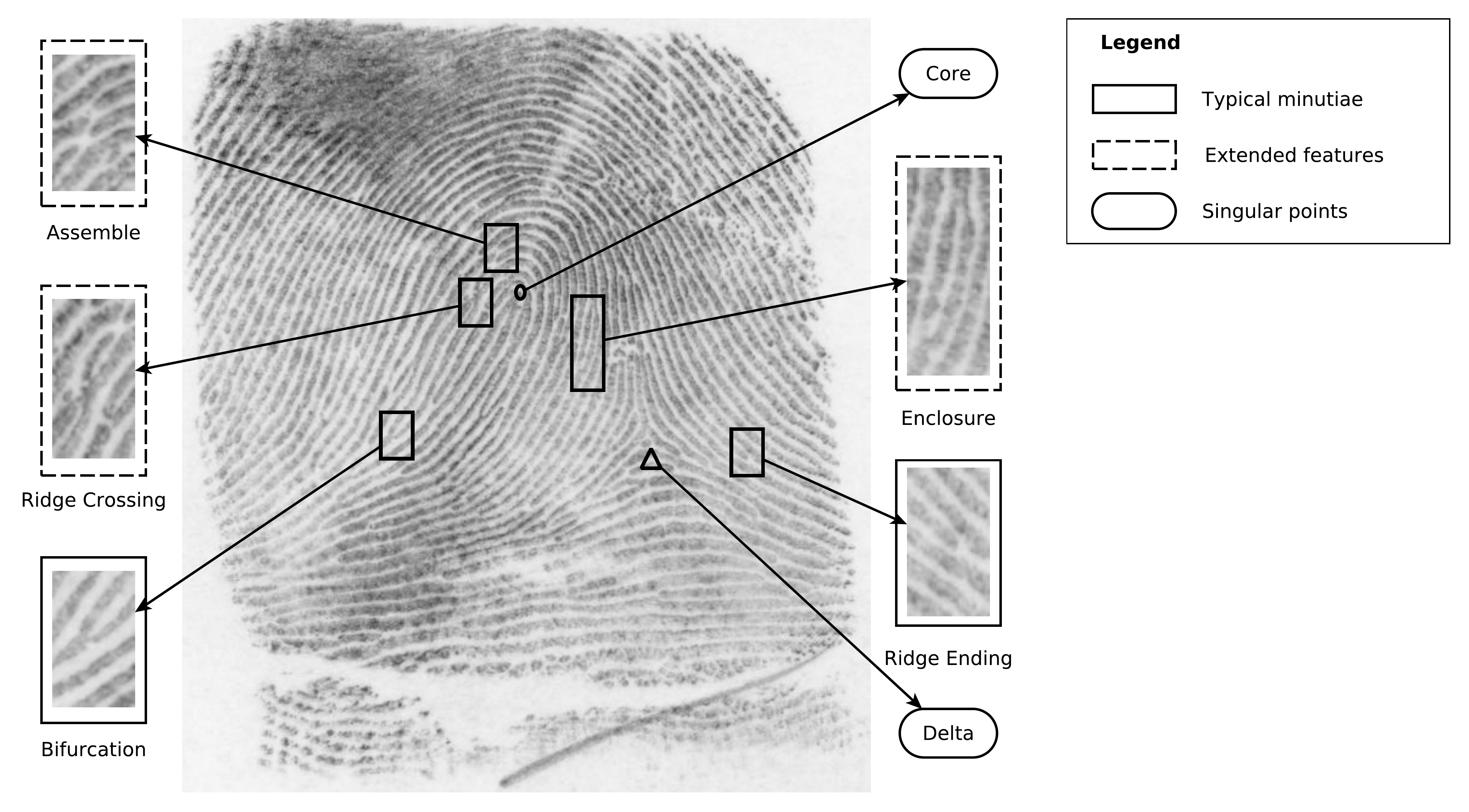}
\caption{Typical minutiae (ridge-ending, bifurcation), extended
  features (assemble, ridge-crossing, enclosure) and singular points
  (core, delta) in an exemplar fingerprint from NIST-SD27 database.}
\label{fig:rareIMG}
\end{figure}

Fingerprint features are categorized into three levels as well as a
feature category called ``other'' to be used for friction ridge
examination. Level-One considers general overall direction
of the ridge flow. Level-Two describes the path of specific
ridges. Level-Three  are the shapes of the ridge
structure. ``Other'' features describe temporary features or
imperfections in ridges~\citep{tfs2011}. Some of the extended fingerprint
features defined by CDEFFS under each of the three level
categories~\citep{stdEFS}~\citep{efsList} are summarized
in Table~\ref{tab:levelEFS}. Figure~\ref{fig:rareIMG} shows some
extended features and typical minutia features (ridge-endings and
bifurcations) in an exemplar fingerprint from NIST Special Database 27
(NIST-SD27).

\begin{table}[t]
\footnotesize
\begin{center}
  \begin{tabular}{|p{2.8cm}|p{6cm}|}
\hline
\vspace{0.005in}\bf{\hspace{0.1in}Type of category}
& \vspace{0.005in}\bf{\hspace{0.5in}Extended feature set}\\
\hline\hline
\vspace{0.1in} Level-One Details &
\vspace{0.005in} Ridge flow map, local ridge quality, pattern
classification (whorl, arch, tentarch, left/right loop etc), singular
points (core, delta), core-delta ridge count. \\
\hline
\vspace{0.1in} Level-Two Details &
\vspace{0.005in} Minutiae-ridge relationship, ridge curvature, feature
relationship, unusual/rare minutiae, scars, creases, incipient ridges,
dots. \\
\hline
\vspace{0.0in} Level-Three Details &
\vspace{0.005in} Pores, edge shapes, ridge/furrow width. \\
\hline
\end{tabular}
\end{center}
\caption{Extended features defined by CDEFFS categorized into
  respective fingerprint feature details (not a comprehensive list).}
\label{tab:levelEFS}
\end{table}

In this work, we propose a method to improve the identification
accuracy of minutiae-based matchers for partial latent fingerprints by
incorporating reliably extracted rare minutia features. Most
minutiae-based fingerprint
matchers use only two prominent ridge characteristics namely
\emph{ridge-endings} and \emph{bifurcations}. We propose an algorithm
that will modify the similarity scores of minutiae-based matchers based on
the presence of rare minutia features like \emph{fragments},
\emph{enclosures}, \emph{dots},  \emph{interruptions, etc}.
The decision for a match or non-match is automatically
estimated based on least squares fitting of an affine
transformation between the latent minutiae set and the tenprint
minutiae set. We show a
significant improvement in the overall rank identification accuracies
for three minutiae-based matchers (NIST-Bozorth3, VeriFinger-SDK and
MCC-SDK) when their similarity scores are modified using our proposed
algorithm which incorporates rare minutia features.
A preliminary version of this work~\citep{rpk2015WIFS} modifies the scores
based on the probability of occurrence of rare minutiae
features. Here, we propose a method which avoids dependencies on such
probability of occurrence as they may vary depending on the size of
the database.

The main contributions of this work are as follows:
\begin{enumerate}

  \item A methodology to adapt any minutiae-based matcher by
    incorporating information from rare features.

  \item A specific algorithm to align the latent minutiae pattern and
    the tenprint minutiae pattern using rare minutiae.

  \item Experimental demonstration of the performance
    improvement of minutiae-based matchers when incorporating
    information from rare features.

  \item We finally present also various population statistics about rare minutia
    features present in a realistic forensic casework database
    obtained from Spanish law enforcement agency (Guardia Civil).

\end{enumerate}

In the following sections, we review related works in the
individualization of fingerprints, and describe: the database and
statistics of rare minutia features, the proposed algorithm to modify
the similarity scores based on rare features, experiments, results and
conclusions.

%-------------------------------------------------------------------------

\section{Related works}
\label{sec:relatedWorks}

To use EFS in automated systems, reliable feature extraction
algorithms are mandatory. Many law enforcement agencies follow
a $500$ ppi scanning resolution for fingerprint images to be used in
AFIS. With such a
resolution, it is difficult to reliably extract many of the extended
features automatically. Due to advances in fingerprint scanning
technologies, SWGFAST during the ANSI/NIST Fingerprint
Standard Update Workshop in $2005$ proposed $1000$ ppi as the minimum scanning
resolution for fingerprint images. This proposal was hugely supported
by the forensic community. To test the feasibility of including EFS in
latent AFIS, NIST conducted another multi-phase commercial
algorithm evaluation called Evaluation of Latent Fingerprint
Technologies - Extended Feature Sets (ELFT-EFS)~\citep{elft-efs}.

ELFT-EFS was conducted in a ``Semi Lights-Out'' mode,
unlike the ``Lights-Out'' mode for ELFT. The main
purpose of ELFT-EFS
was to determine the effectiveness of forensic examiner marked latent
fingerprint features on the latent identification accuracy. NIST
conducted two evaluations for ELFT-EFS and the best achieved Rank-1
identification accuracy for each of the evaluations is summarized in
Table~\ref{tab:efsRank}.

\begin{table}[t]
\footnotesize
\begin{center}
\begin{tabular}{|p{3.2cm}|p{4cm}|p{1.5cm}|}
\hline
\vspace{0.005in}\bf{\hspace{0.3in} ELFT-EFS}
& \vspace{0.005in}\bf{\hspace{0.5in}Database size} &
\bf{Rank-1\newline accuracy}\\
\hline\hline
\vspace{0.1in} Evaluation-1 \cite{elft-efs} & \vspace{0.005in} 1,114 latents
compared against 1,000,000 rolled prints and 1,000,000 plain prints
& \vspace{0.1in}$66.7\%$ \\
\hline
\vspace{0.1in} Evaluation-2 \cite{elft-efs2}~\cite{elft-efs22} &
\vspace{0.005in} 1,066 latents
compared against 1,000,000 rolled prints and 1,000,000 plain prints
& \vspace{0.1in}$71.4\%$ \\
\hline
\end{tabular}
\end{center}
\caption{Rank-1 identification accuracy of NIST Evaluation of Latent
  Fingerprint Technologies - Extended Feature Sets (ELFT-EFS).}
\label{tab:efsRank}
\end{table}

As in ELFT, the results of different evaluations in ELFT-EFS cannot be
directly compared because the database used were not exactly the
same~\citep{elft-efs2}~\citep{elft-efs}. In~\citep{elft-efs2}, it is
reported that though the highest measured accuracy achieved by a
individual matcher at Rank-1 was $71.4\%$, and approximately $82\%$ of the
latents were correctly matched at Rank-1 when
multiple matchers were combined. This
corroborates the potential for additional accuracy improvement when
combining multiple
algorithms~\citep{Fierrez-Aguilar2006_FusionFingerprintQ}. Nevertheless,
these NIST evaluations show that the performance of state of the art
latent fingerprint recognition technologies are not satisfactory.

An extensive study on extended fingerprint feature sets is reported
by Jain~\citep{jainEFS}. This includes several extended features from
Level-One,
Level-Two and Level-Three. It was concluded
in~\citep{jainEFS} that manual intervention is strongly recommended
while using EFS, as well as extended features from Level-One and
Level-Two are highly recommended to be incorporated in latent
AFIS. Extended features such as ridge flow map, ridge wavelength
map, ridge quality map, and ridge skeleton have shown significant
improvements in latent identification accuracies. Level-One and
Level-Two details used
in~\citep{jainEFS}~\citep{jain2011latent} are insensitive to image quality,
and do not rely on high resolution images. To incorporate Level-Three
EFS such as pores, dots, incipients, etc, it is essential to improve the
quality of enrolled fingerprints.

The use of pores as extended features was studied in high
resolution $1000$ ppi images by Zhao et al.~\citep{zhaoEFS} and Jain
et al.~\citep{pores}. Dots and
incipients were studied by Chen et al.~\citep{dots}. Out of pores, dots and
incipients, pores resulted in better
performance~\citep{zhaoEFS}. Even
though high resolution $1000$ ppi images were used, live scan
images resulted in easy detection of pores automatically,
which was not the case with inked fingerprint images. Pore
extraction based on skeletonized and binary images was studied
by Stosz et al.~\citep{stosz, stosz2} and Kryszczuk et al.~\citep{kry}. These
techniques were demonstrated
effective only on very good quality high resolution fingerprint images scanned
approximately at $2000$ ppi~\citep{stosz}. These methods were more
sensitive to noise, and the performance degrades for poor quality of
fingerprint images and low resolution images.

A local image quality based method applied on extended fingerprint
features for high
resolution $1000$ ppi fingerprint images was reported
by Vatsa et al.~\citep{vatsa}. A fast curve evolution algorithm was
used to quickly
extract extended features such as pores, ridge contours, dots and
incipient ridges. Together with other Level-One and Level-Two details
as proposed in Jain et al.~\citep{jain2011latent},
these extended features were used to generate a quality-based
likelihood ratio to improve the identification performance.

Score level fusion of different algorithms using various extended
fingerprint features was reported by Fierrez et al.~\citep{fierrez}. Features like
singular points, ridge skeleton, ridge counts, ridge flow map, ridge
wavelength map, texture measures were studied by analyzing the
correlation between them using feature subset-selection
techniques. Combination of features showed significant improvement in
the performance of the system.

When only partial fingerprints are available, pre-alignment
of partial minutiae set and full minutiae set based on orientation
fields of respective fingerprints helps in reducing
the minutiae search space of full fingerprint relative to the
size of partial fingerprint. Such reduction in the size of
minutiae search space has been shown to improve the performance by
Krish et
al.~\citep{rpkIET},~\citep{rpk2014ICPR},~\citep{rpk2014IWBF}. This
approach has shown significant improvement in the system performance
especially for poor quality latent fingerprints.

Si et al.~\citep{SI201787} combined local and global approaches for
minutiae matching. Their proposed method estimates a dense deformation
field between two fingerprints to remove the negative impact of
distortion. The dense deformation field aligns not only minutiae but
also ridges. By fusing minutiae and image correlation, they improved
the matching performance significantly.

Cao and Jain~\citep{caojain2017} proposed the use of
minutiae descriptors that are learned via a ConvNet together with
minutiae and texture information to improve the latent fingerprint
recognition. They performed score level fusion of their proposed
algorithm with commercial minutiae-based matchers to improve the rank
identification accuracies. We followed a similar methodology to
show the importance of rare minutiae features in improving the rank
identification accuracies of minutiae-based matchers.

%-------------------------------------------------------------------------

\section{Database}
\label{sec:database}

Similar to the related works discussed in Sect. 1, we first tried to use SD27 data from NIST in our experiments. Regretfully, the public distribution of SD27 data from NIST is now discontinued. Additionally, copies already distributed of SD27 lack ground truth information of the existing rare features, therefore we decided to generate and make public a new dataset similar to SD27 but in this case including ground truth information generated by forensic experts.\footnote{The dataset is available here: \url{https://atvs.ii.uam.es/atvs/gcdb_features.html}}

\begin{figure}[t]
\centering
\includegraphics[height=3cm]{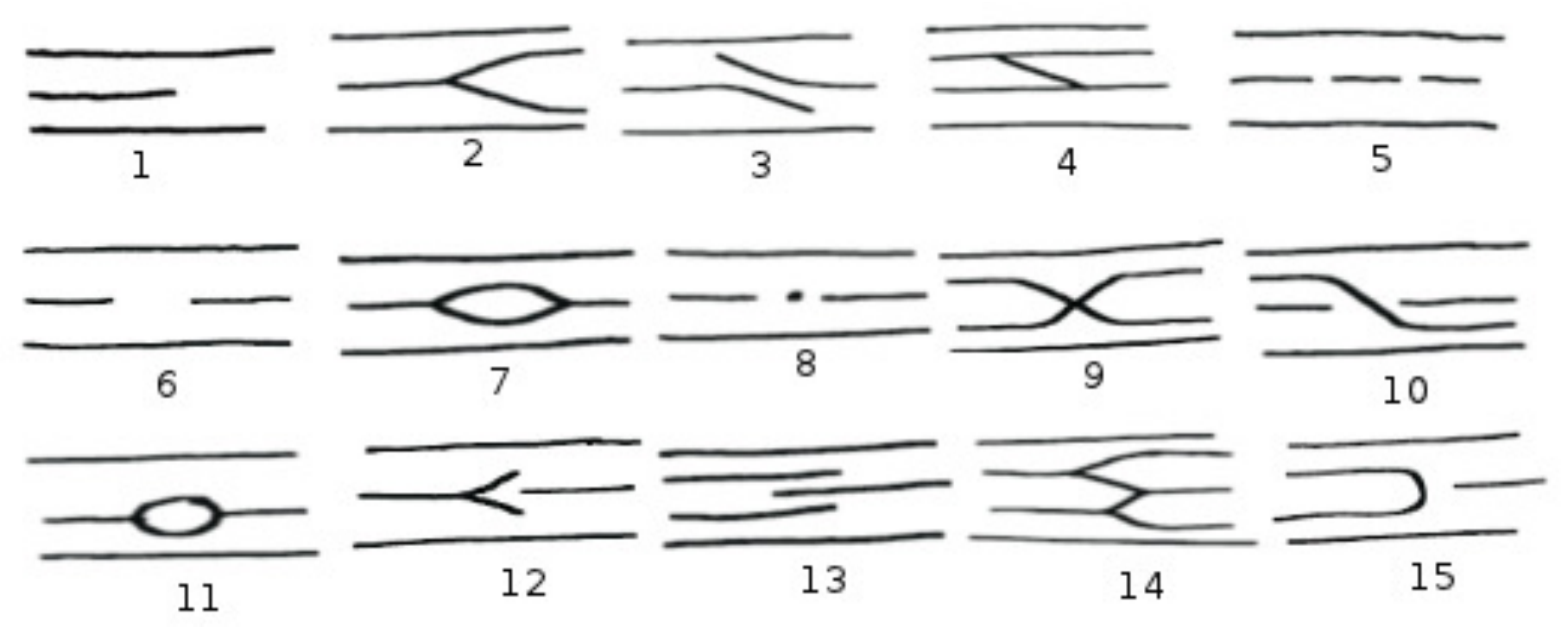}
\caption{Minutia types used by Guardia Civil. Names
      corresponding to individual minutia type numbers can be found in
      Table~\ref{tab:gcListTable}.}
\label{fig:gcList}
\end{figure}

The database used in this work was obtained from Guardia Civil, the
Spanish law enforcement agency. The Guardia Civil database (GCDB) is a
realistic forensic fingerprint casework database.
Apart from having typical minutiae feature types
(\emph{ridge-endings, bifurcations}), GCDB also comprises rare
minutiae types like \emph{fragments}, \emph{enclosures},
\emph{dots},  \emph{interruptions, etc}~\cite{GC01}. A comprehensive
list of rare minutia features used by
Guardia Civil are shown in
Figure~\ref{fig:gcList} and the corresponding minutiae type names are
listed in Table ~\ref{tab:gcListTable}.

GCDB used in this work consists of 268 latent and tenprint (exemplar)
pairs of fingerprint images (scanned at $500$ ppi) and minutia sets. All
the minutiae in the
latent fingerprint images were manually extracted by forensic
examiners of Guardia Civil. The corresponding mated minutiae in the
tenprints were also manually established. This includes the typical
(ridge-endings and bifurcations) minutiae and the rare minutiae.
These are called \emph{matched} minutiae set, i.e, the minutiae
sets for which a one-to-one correspondence is established between the
latent and the mated tenprint. Here, the number of minutiae in
the latent and the corresponding mated tenprint minutiae set are the
same.

\begin{table}[t]
\footnotesize
\begin{center}
\renewcommand{\arraystretch}{1.4}
\setlength\tabcolsep{3pt}
\begin{tabular}{|c|c|c|c|c|c|}
\hline\noalign{\smallskip}
\bf{No} & \bf{Minutiae type} & \bf{No} & \bf{Minutiae type} & \bf{No}
& \bf{Minutiae type} \\
\noalign{\smallskip}
\hline
\noalign{\smallskip}
1 & Ridge Ending & 6 & Interruption & 11 & Circle\\
2 & Bifurcation & 7 & Enclosure & 12 & Delta\\
3 & Deviation & 8 & Point & 13 & Assemble\\
4 & Bridge & 9 & Ridge Crossing & 14 & M-structure\\
5 & Fragment  & 10 & Transversal & 15 & Return\\
\hline
\end{tabular}
\end{center}
\caption{List of minutia types used by Guardia Civil. Numbering with respect to
  Figure \ref{fig:gcList}.}
\label{tab:gcListTable}
\end{table}

To generate an \emph{ideal} minutiae set (i.e, all possible minutiae)
for the tenprint, we used the minutiae extractor module from
VeriFinger-SDK~\cite{cVeriFinger}. We performed a Gabor filtering
based global post-processing to remove any spurious minutiae that are outside
the foreground. In particular, Gabor filtering is used to
obtain the Region of Interest (RoI) similarly
to~\cite{Alonso-Fernandez2005_ISPA_FingerprintSegmentation}.
The spurious minutiae generated by VeriFinger outside the RoI are then
eliminated. VeriFinger extracts only the typical minutiae
featuresfrom the fingerprint image. We then added the rare minutiae
from the GCDB tenprint minutiae set into the post-processed VeriFinger generated
minutiae set for the tenprints. In this case the number of minutiae
between the latent and the corresponding mated tenprint minutiae set
are not equal, the latent minutiae set is only a subset of the tenprint
minutiae set. The average number of minutiae in
the latents was $13$ and that of tenprints was $125$. There exists
automated algorithms to perform segmentation for latent
fingerprints~\citep{anush03}, but in this work, we relied on
the manually extracted minutiae alone.

\begin{table}[t]
\footnotesize
\begin{center}
  \begin{tabular}{|p{0.6cm}|p{2cm}|p{2cm}|p{1.8cm}|}
\hline
\vspace{0.005in}\bf{\hspace{0.04in}No}\hspace{0.0in} &
\vspace{0.005in}\bf{\hspace{0.0in}Minutiae Type} &
\vspace{0.005in}\bf{\hspace{0.0in}Probability ($p_i$)} &
\vspace{0.005in}\bf{\hspace{0.0in}Number of occurrences}\\
\hline
\vspace{0.005in}\hspace{0.1in}1 & \vspace{0.005in}\ \ Ridge-ending &
\vspace{0.005in}\hspace{0.2in}\ \ 0.5634 & \vspace{0.005in}\hspace{0.2in}\ \ 1902\\

\vspace{0.005in}\hspace{0.1in}2 & \vspace{0.005in}\ \ Bifurcation &
\vspace{0.005in}\hspace{0.2in}\ \ 0.3620 & \vspace{0.005in}\hspace{0.2in}\ \ 1222\\

\vspace{0.005in}\hspace{0.1in}3 & \vspace{0.005in}\ \ Deviation &
\vspace{0.005in}\hspace{0.2in}\ \ 0.0015 & \vspace{0.005in}\hspace{0.2in}\ \ 5\\

\vspace{0.005in}\hspace{0.1in}4 & \vspace{0.005in}\ \ Bridge &
\vspace{0.005in}\hspace{0.2in}\ \ 0.0024 & \vspace{0.005in}\hspace{0.2in}\ \ 8\\

\vspace{0.005in}\hspace{0.1in}5 & \vspace{0.005in}\ \ Fragment &
\vspace{0.005in}\hspace{0.2in}\ \ 0.0444 & \vspace{0.005in}\hspace{0.2in}\ \ 150\\

\vspace{0.005in}\hspace{0.1in}6 & \vspace{0.005in}\ \ Interruption &
\vspace{0.005in}\hspace{0.2in}\ \ 0.0021 & \vspace{0.005in}\hspace{0.2in}\ \ 7\\

\vspace{0.005in}\hspace{0.1in}7 & \vspace{0.005in}\ \ Enclosure &
\vspace{0.005in}\hspace{0.2in}\ \ 0.0204 & \vspace{0.005in}\hspace{0.2in}\ \ 69\\

\vspace{0.005in}\hspace{0.1in}8 & \vspace{0.005in}\ \ Point &
\vspace{0.005in}\hspace{0.2in}\ \ 0.0036 & \vspace{0.005in}\hspace{0.2in}\ \ 12\\

\vspace{0.005in}\hspace{0.1in}10 & \vspace{0.005in}\ \ Transversal &
\vspace{0.005in}\hspace{0.2in}\ \ 0.0003 & \vspace{0.005in}\hspace{0.2in}\ \ 1\\

\hline

\end{tabular}
\end{center}
\caption{The probability of occurrence of minutia types present
  in the $268$ latent fingerprints of GCDB (total number of minutiae
  observed = 3376). The numbers correspond to minutia types in
  Figure~\ref{fig:gcList}}
\label{tab:gcStat}
\end{table}

The original latent minutia sets provided by Guardia
Civil and the post-processed VeriFinger generated minutia sets are
used in all our experiments. To represent some rare minutiae, multiple
points were needed. For example, to represent a \emph{deviation} two
points are needed (see type 3 in Figure~\ref{fig:gcList}), and to
represent an \emph{assemble} three points are
needed (see type 13 in Figure~\ref{fig:gcList}).
Whenever multiple
points are needed to represent a rare
minutia, we mapped them to a single point representation by taking the
average of locations of all points, and minimum orientation among all
the orientations.

From the $268$ latent fingerprint minutia sets, we estimated the
probability of occurrence ($p_i$) of various minutia types. The
probability ($p_i$) for each minutia type present in GCDB is listed in
Table~\ref{tab:gcStat}.
In the $268$ latent fingerprints of GCDB, we noticed only seven types of
rare minutia features. They are listed in
Table~\ref{tab:gcStat}. Other rare minutia types are not found in the
current database used in this study. This is
  particularly due to the fact that we are dealing with highly partial
latent fingerprints, with an average of $13$ minutiae per latent.

In related works, minutiae frequency for $20$ different minutiae types were reported for full fingerprints obtained from a population of $200$ Spanish individuals~\citep{GUTIERREZREDOMERO2012266} and $278$ Argentinian individuals~\citep{GUTIERREZREDOMERO201179}. The statistics of the various minutiae types obtained in those works (for ca. $2,000$ different fingerprints, more than $80,000$ minutiae) are similar in many cases to the statistics obtained in the partial latent fingerprint database presented in the present work if we compare corresponding types of features, especially for the most frequent minutiae types. For the comparison, please note that the naming conventions differ among works. Here we follow the naming conventions of Guardia Civil, e.g., Type 6 in Fig. 3 is called ``Interruption'', whereas this kind of feature is called ``Break (BR)'' in works~\citep{GUTIERREZREDOMERO2012266} and~\citep{GUTIERREZREDOMERO201179}. Comparing for example our statistics from Table~\ref{tab:gcStat} to Table 1 of~\citep{GUTIERREZREDOMERO2012266}, and using approximated values, our Type 1 is 56\% vs Type E in~\citep{GUTIERREZREDOMERO2012266} is 50\%, our Type 2 is 36\% vs Type B+C is 40\%, our Type 7 is 2\% vs Type ENBG+ENSM is 2\%, etc. A few other types, specially the rarest, have larger differences in their statistics, most probably because of the sample size, e.g., our Type 3 is ca. 0.1\% vs Type O in~\citep{GUTIERREZREDOMERO2012266} is 0.7\%.

Given our limited sample size, please note that the statistics from Table~\ref{tab:gcStat} are not pretending to be representative of large-scale populations, especially for the rarest features. It would be necessary a really large dataset to generate statistically significant frequencies for all the reported types, which is out of the scope of the present paper.

%-------------------------------------------------------------------------

\section{Algorithm}
\label{sec:algoFE}

We assume in our algorithm that the rare features won't be always
repeatable and won't be always labelled uniformly over multiple
captures (either manually or automatically), e.g., due to acquisition
variations or other image quality
factors~\cite{Alonso-Fernandez2007-FingerQualityReview,2011-QualityBio-FAlonso}. Similarly to
minutiae matchers able to cope with variations between matching
fingerprint images, our algorithm is also able to cope with such
intra-variability. A comprehensive study of such variability factors
is out of the scope of the present work. Here we only focus in
analyzing to what extent those rare minutiae can improve existing
AFIS, using realistic data in which such intra-variations between
multiple captures are naturally present.
The latent fingerprints of GCDB are highly partial in nature, with an
average of $13$ minutiae per latent. To make an appropriate alignment
between the latent minutia points and the tenprint minutia
points (with an average of $125$ minutia points) requires a reliable
reference point. We choose the rare minutia
features as reference points to perform the alignment.

Let $L$ and $M$ be the representation of latent and
tenprint minutia sets respectively. Each minutia is represented as a
quadruple $m = \{x,y,\theta, t\}$ that indicates the $(x,y)$ location
as coordinates, the minutia angle $\theta$ and the minutia type $t$:
\begin{equation*}
  L = \left[m_1\ m_2\ ...\ m_p\right],
  \hspace{0.1in} m_i = [x_i\  y_i\  \theta_i\  t_i]^T,
  \hspace{0.1in}i = 1...p
\end{equation*}
\begin{equation*}
  M = [m'_1\  m'_2\ ...\ m'_q],
  \hspace{0.1in} m'_j = [x'_j\  y'_j\  \theta'_j\  t'_j]^T,
  \hspace{0.1in}j = 1...q,
\end{equation*}
where $p$ and $q$ are the number of minutiae in $L$ and $M$
respectively. If $t > 2$, then the minutia is of rare type (from
Table~\ref{tab:gcListTable}), and $[\ \cdot\ ]^T$ denotes transpose.

The algorithm to generate weighted similarity scores from a minutiae
matcher is described in two stages. Similarity scores of minutiae
matchers are modified only if they contain rare minutia features.

\begin{figure}[t]
\centering
\includegraphics[scale=0.5]{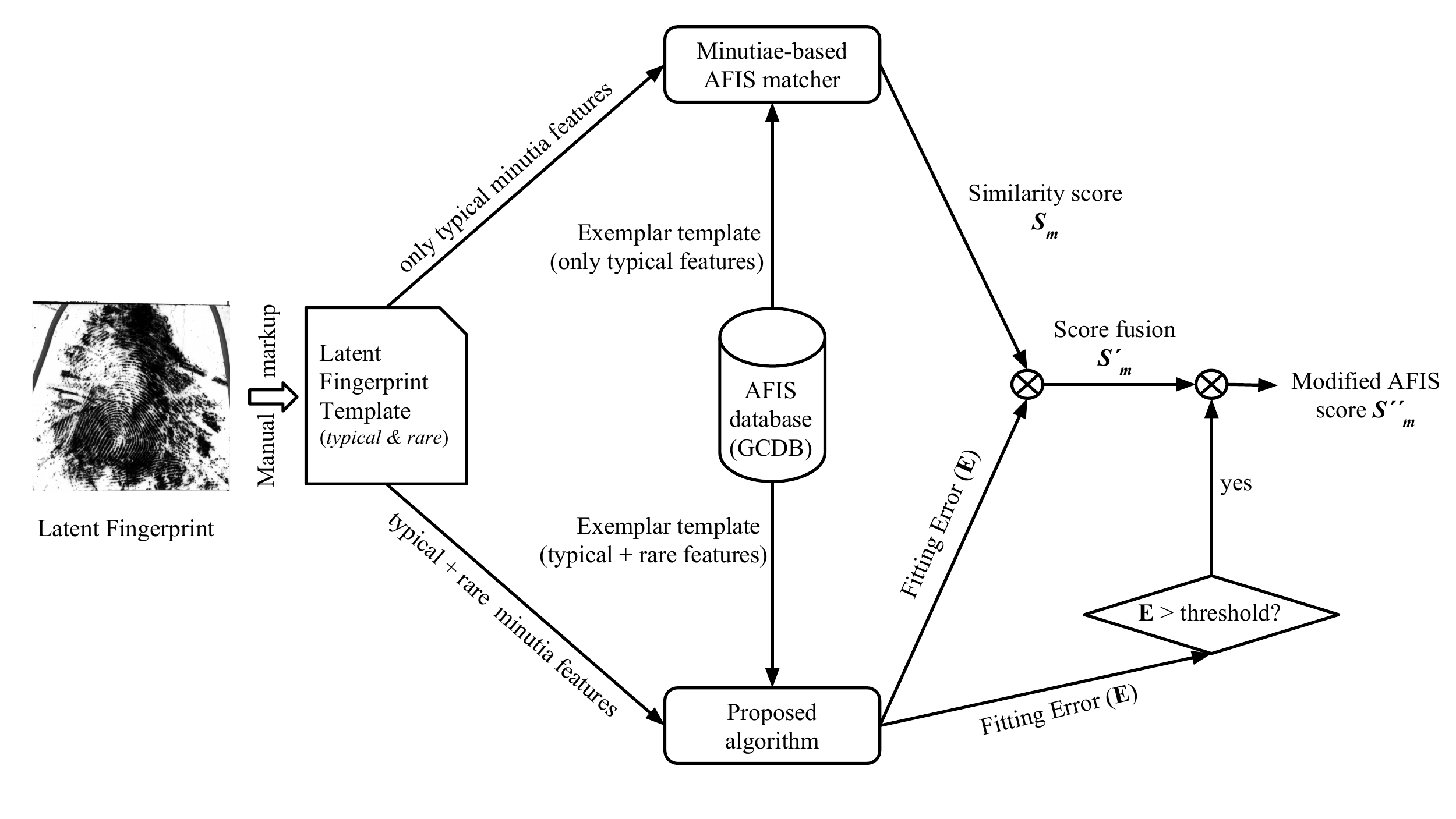}
\caption{Sequence of steps in estimating the modified similarity score
  of a reference minutiae-based matcher. Note that the AFIS matcher
  that we want to improve runs in parallel and independently of our
  proposed algorithm based on rare features. Our proposed algorithm
  generates a new score that is fused to the standard output score
  provided by the AFIS matcher.}
\label{fig:algo}
\end{figure}

The first stage of the algorithm
estimates the least square fitting error for an affine transformation
of the latent minutiae set onto a tenprint minutiae set.
The second stage of the algorithm modifies the similarity score
generated by the minutiae-based matcher based on the fitting
error. Other works related with modifying the similarity score based
on pre-alignment are reported in~\citep{paulinoMCC, rpkIET, msccICB2015}.
The sequence of steps involved in generating the modified score of the
minutiae matcher using our proposed algorithm is summarized in
Figure~\ref{fig:algo}.

\subsection*{\bf\emph{Stage-1 : Least Square Fitting Error}}
\vspace{0.1in}
{\bf{Step 1}}: To find the affine transformation between $L$ and $M$, it is
first needed to establish a one-to-one correspondence between
minutiae from $L$ and minutiae from $M$. Let the
subset of minutiae from $M$ which establishes correspondence with $L$
be denoted as $M_s$.

{\bf{Step 2}}: Superimpose one rare minutia point of $L$ onto the
corresponding rare minutia point of $M$, only if
they both are of the same type (if there are multiple rare minutia
points, take any).
This step compensates the translation between latent and tenprint. If
the type of the rare minutia between $L$ and $M$ differs, or $M$ does not
contain any rare minutiae, then the similarity score generated by our
algorithm is fixed to a value 0.25 inside a range (0,1). That
similarity score will then be combined with the score provided by the
AFIS system (see Figure~\ref{fig:algo}). Note that we align based on rare minutiae of
the same type between latent and tenprint. Provided that rare minutia
are more unique than standard minutia and that the baseline AFIS that
we are trying to improve most probably won't be exploiting such
specific rare types, we expect that our combination approach shown in
Figure~\ref{fig:algo} will fuse complementary information resulting in improved
performance.

%%\vspace{0.2in}
{\bf{Step 3}}: To establish the correspondence between latent and tenprint
minutia points, we choose the minutia points from $M$
that are close to the minutia points of $L$. The Euclidean distance is
calculated between the minutia pairs to determine whether the pairs
are close or not.

{\bf{Step 4}}: To compensate for rotation alignment, we rotate the
latent in the range $[-45^\circ, +45^\circ]$ with respect to the
superimposed rare minutiae, and estimate the Euclidean distance for
each rotation step of size $1^\circ$.

{\bf{Step 5}}: The optimal rotation is the one for which the
average sum of distances between closest pairs is minimum.

{\bf{Step 6}}: After the alignment, all those minutia pairs which are
within a threshold distance are considered to be mated pairs, and a
one-to-one correspondence is established between them.  As a result,
we obtain a subset $M_s$ of the tenprint minutiae $M$.
After establishing the correspondence, the number of
minutiae between $L$ and $M_s$ are the same.

{\bf{Step 7}}: Once the correspondence is established, we find the
least square fitting error for the affine transformation between the latent
minutia points and the subset of tenprint minutiae set.

For $\hat{L}$ and $\hat{M_s}$, which are the modified version of $L$
and $M_s$ with only the $(x,y)$ locations as minutia representation
augmented with a value $1$, i.e,:

\begin{equation*}
  \hat{L} = [\hat{m_1}\  \hat{m_2}\ ...\ \hat{m_p}];
   \hspace{0.1in} \hat{m_i} = [x_i\ y_i\ 1]^T;
  \hspace{0.1in} i = 1...p
\end{equation*}
\begin{equation*}
  \hat{M_s} = [\hat{m'_1}\  \hat{m'_2}\ ...\ \hat{m'_p}];
  \hspace{0.1in} \hat{m'_j} = [{x'_j}\ {y'_j}\ 1]^T;
  \hspace{0.1in} j = 1...p,
\end{equation*}

we are looking for some affine transformation matrix \vspace{0.2in}
\begin{equation}
  A = [a_{jk}]_{j,k=1...3}
\end{equation}
and some translation vector %\vspace{0.2in}
\begin{equation}
  \tau = [\tau_1\  \tau_2\ ...\ \tau_p];
  \hspace{0.1in} \tau_1 = \tau_2 = ... = \tau_p =
         [\delta_x\ \delta_y\ 1]^T;
\end{equation}
such that %\vspace{0.2in}
\begin{equation}
  \hat{M_s} \approx A \hat{L} + \tau
\end{equation}
where $[\delta_x\ \delta_y]$ is the translation needed to superimpose
the rare minutia of latent minutia set $L$ and tenprint minutia set
$M$.\\

{\bf{Step 8}}: Find the least square fitting error between $\hat{L}$
and $\hat{M_s}$  defined as follows:
\begin{equation}
  E^{\hat{L},\hat{M_s}} = \frac{1}{p} \sum_{i = 1}^{p}||\hat{m'_i} - A
  \hat{m_i}  - \tau_i||_2^2
\end{equation}
where $||\hat{m'_i} - A\hat{m_i}  - \tau_i||_2$ is the $L_2$ norm.

For a match comparison, we expect this fitting error to be small,
whereas for a non-match comparison, the fitting error is expected to
be large.

If there are multiple matching rare minutiae feature between $L$ and
$M_s$, then $E^{\hat{L},\hat{M_s}}$ is calculated
for all such minutiae types. The fitting error for such a comparison
is chosen to be the minimum of all the fitting errors calculated.

\subsection*{\bf\emph{Stage-2 : Modified scores}}

\vspace{0.1in}
{\bf{Step 9}}: Using a standard minutia matcher, generate the
similarity score $S_m$ between $L$ and $M$. Assuming that the
similarity score values generated by the minutiae matcher is
normalized in range $[0, 1]$. If by default the minutiae-based
matcher do not generate normalized scores, then they are explicitly
normalized.

The fitting error $E^{\hat{L},\hat{M_s}}$ is a dissimilarity
score, i.e, lower the fitting error, comparison is more similar. The
fitting error is normalized in the range $[0, 1]$ and then they are
converted into a similarity score $\hat{E}^{\hat{L},\hat{M_s}}$ as
follows:

\begin{equation}
\hat{E}^{\hat{L},\hat{M_s}} = 1 - E^{\hat{L},\hat{M_s}}
\label{eq:fittingError}
\end{equation}

$S_m$ and $\hat{E}^{\hat{L},\hat{M_s}}$ are similarity scores in the
range $[0, 1]$. First level of score modification is done by score
fusion of $S_m$ and $\hat{E}^{\hat{L},\hat{M_s}}$ based on mean rule
to obtain $S'_m$ as follows:

\begin{equation}
S'_m = \frac{S_m + \hat{E}^{\hat{L},\hat{M_s}}}{2}
\label{eq:fusion}
\end{equation}

{\bf{Step 10}}:
Further to the score modification based on fusion, similarity
score $S'_m$ is again modified based on a fitting error threshold $E_t$
to obtain $S''_m$ as follows:

\begin{equation}
S''_m =
\begin{cases}
S'_m \times \alpha & \text{if $\hat{E}^{\hat{L},\hat{M_s}} > E_t$},\\
S'_m \times \beta & \text{otherwise.}
\end{cases}
\label{eq:feThreshold}
\end{equation}
where $\alpha$ and $\beta$ are constants used to reward and penalize
the fused score $S'_m$ respectively.

If $\hat{E}^{\hat{L},\hat{M_s}} > E_t$, then the
comparison is deemed to be a match, and the fused score is further
rewarded by multiplying with a constant $\alpha$ to obtain $S''_m$. If
$\hat{E}^{\hat{L}, \hat{M_s}} \le E_t$, the comparison is deemed to be a
non-match, and the fused score $S'_m$ is penalized by multiplying it
with constant $\beta$ to obtain $S''_m$.

Thus, we obtain a modified similarity scores $S''_m$ for
a particular minutiae matcher by incorporating information from rare
minutia features based on the fitting error obtained using our
approach.

%-------------------------------------------------------------------------
\section{Experiments}

GCDB consists of 268 latent and corresponding 268 mated
tenprint images and minutiae. Only 151 of them contains rare minutiae
in the minutiae set, and the remaining consisted only of bifurcations
and ridge-endings.
We performed all our experiments on the 151 latent and tenprint
minutiae set to establish the importance of rare minutiae using our
proposed algorithm in improving the rank identification accuracy of
minutiae-based matchers which uses only typical minutiae. To
generate similarity scores based on typical minutia features, we used
three  minutiae matchers namely:
NIST-Bozorth3~\citep{nistB3}, VeriFinger-SDK~\citep{cVeriFinger} and
MCC-SDK~\citep{mcc1, mcc2, mcc3}.
When reporting the rank identification accuracies in our experiments,
there are $151$ match comparisons and $151 \times 150$ non-match
comparisons.

NIST-Bozorth3 is a minutiae based fingerprint matcher that is
specially developed to deal with latent fingerprints and is
publicly available. This matcher is
part of the NIST Biometric Image Software (NBIS)~\citep{nistB3}, developed
by NIST.  VeriFinger is a commercial SDK
that is widely used in academic research. MCC-SDK is
a well known minutiae matcher made available for research purposes.
Both NIST-Bozorth3 and MCC-SDK are publicly available minutiae
matchers, whereas VeriFinger is not.
We report the performance accuracy and
improvement of all the matchers using Cumulative Match Characteristic
(CMC) curves.

\begin{figure}[t]
\centering
\includegraphics[width=9cm]{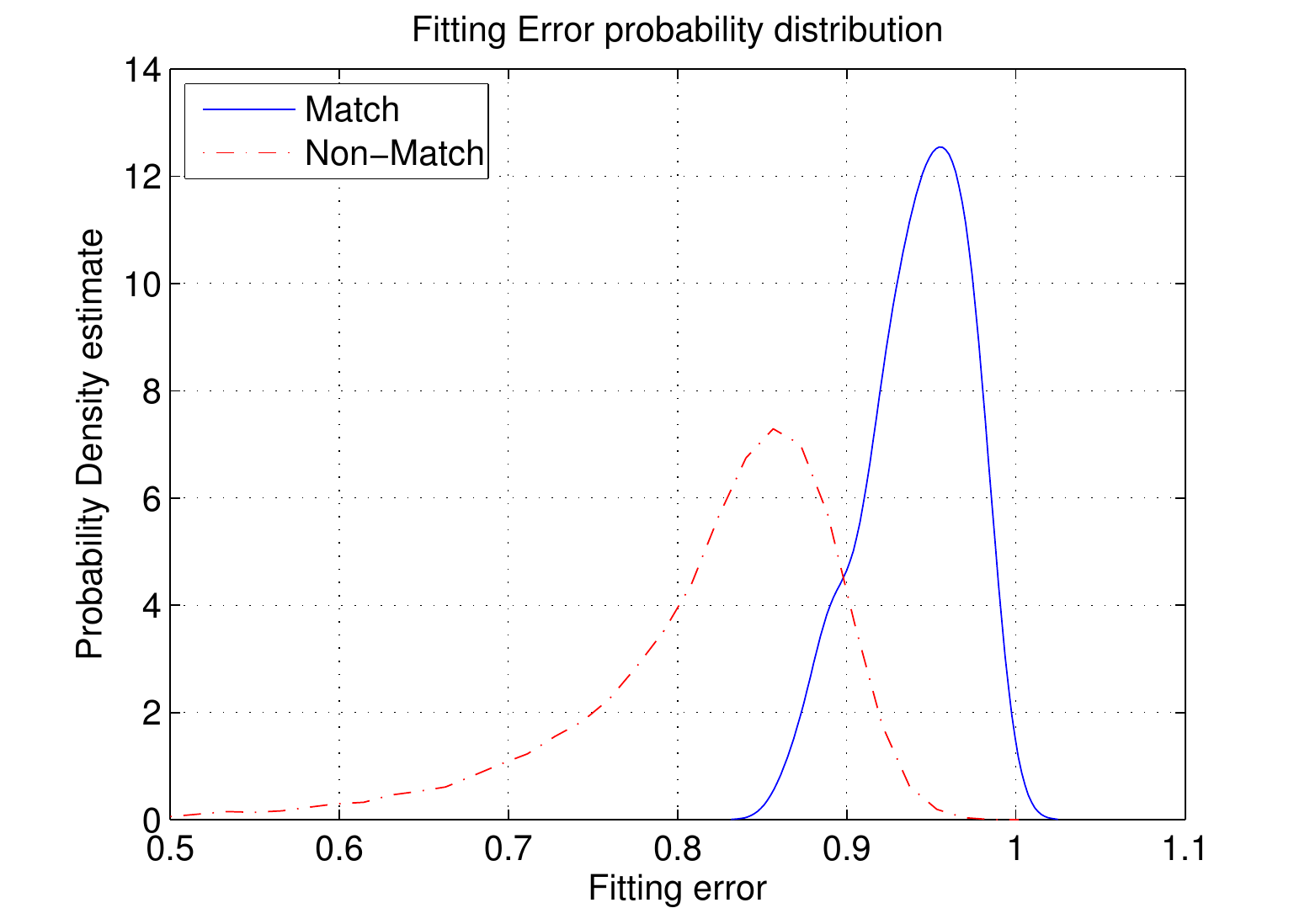}
\caption{Probability density estimate of the fitting errors for match
  and non-match comparisons.}
\label{fig:exp1}
\end{figure}

\subsection{Experiment 1: Fitting Error probability distribution}

The least square fitting error probability density estimates for both
match and non-match comparisons are shown in
Figure~\ref{fig:exp1}. We can observe that the fitting error itself
is discriminatory enough, having separate peaks for both match and
non-match distributions. This supports the methodology followed in our
algorithm. The following experiments also support this fact.

\subsection{Experiment 2: Score Fusion}

The fitting error is a similarity score in the range $[0,1]$ from Eq.
(\ref{eq:fittingError}). The similarity scores obtained from
minutiae-based matchers are normalized in the range $[0,1]$. These
scores are fused based on sum rule (Step 9 of Algorithm).

\begin{figure}[t!]
\begin{center}
  \subfigure[CMC curve for NIST-Bozorth3 before and after score fusion]
            {\includegraphics[width=6.3cm]{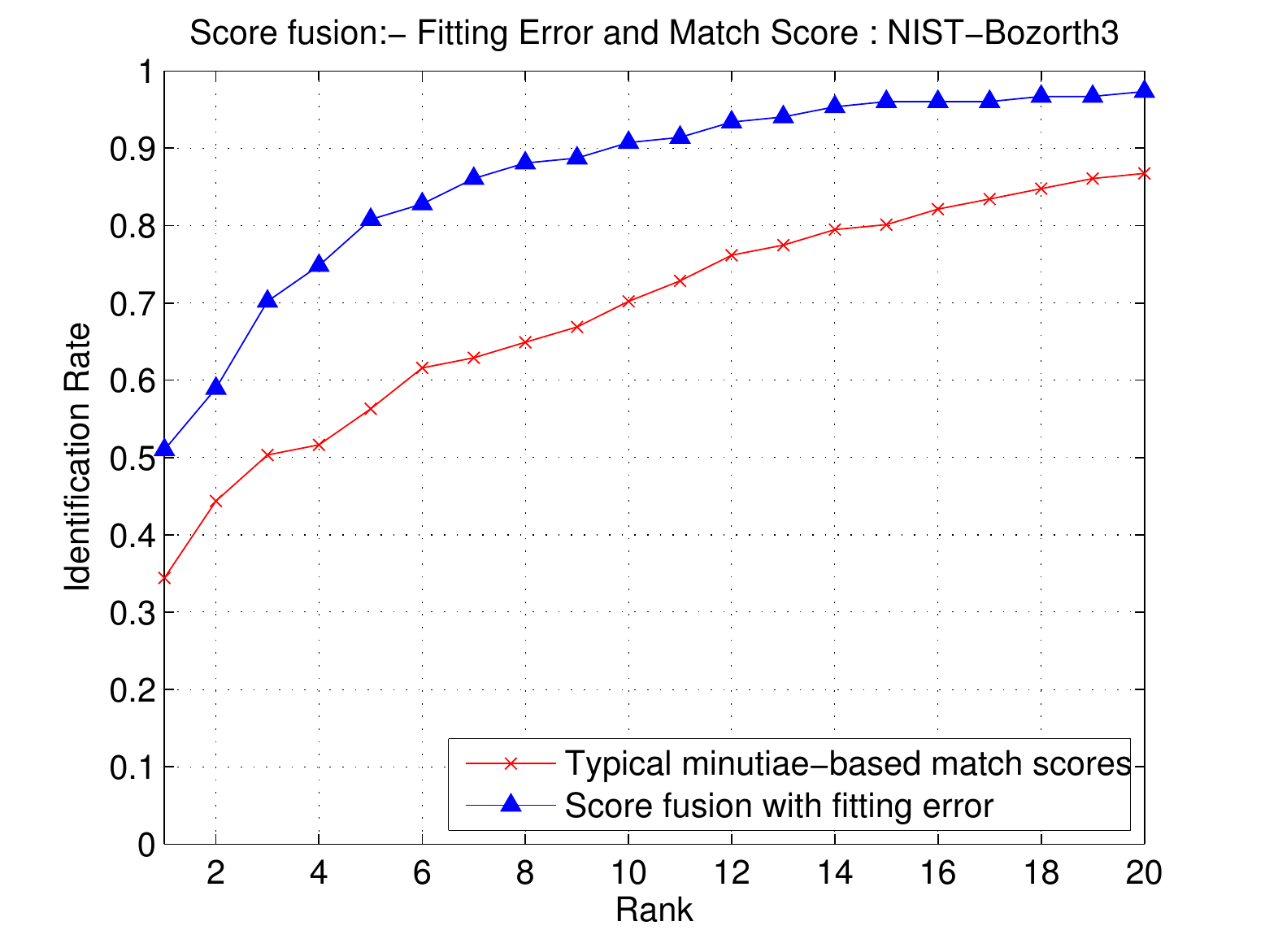}}
  \subfigure[CMC curve for VeriFinger-SDK before and after score fusion]
            {\includegraphics[width=6cm]{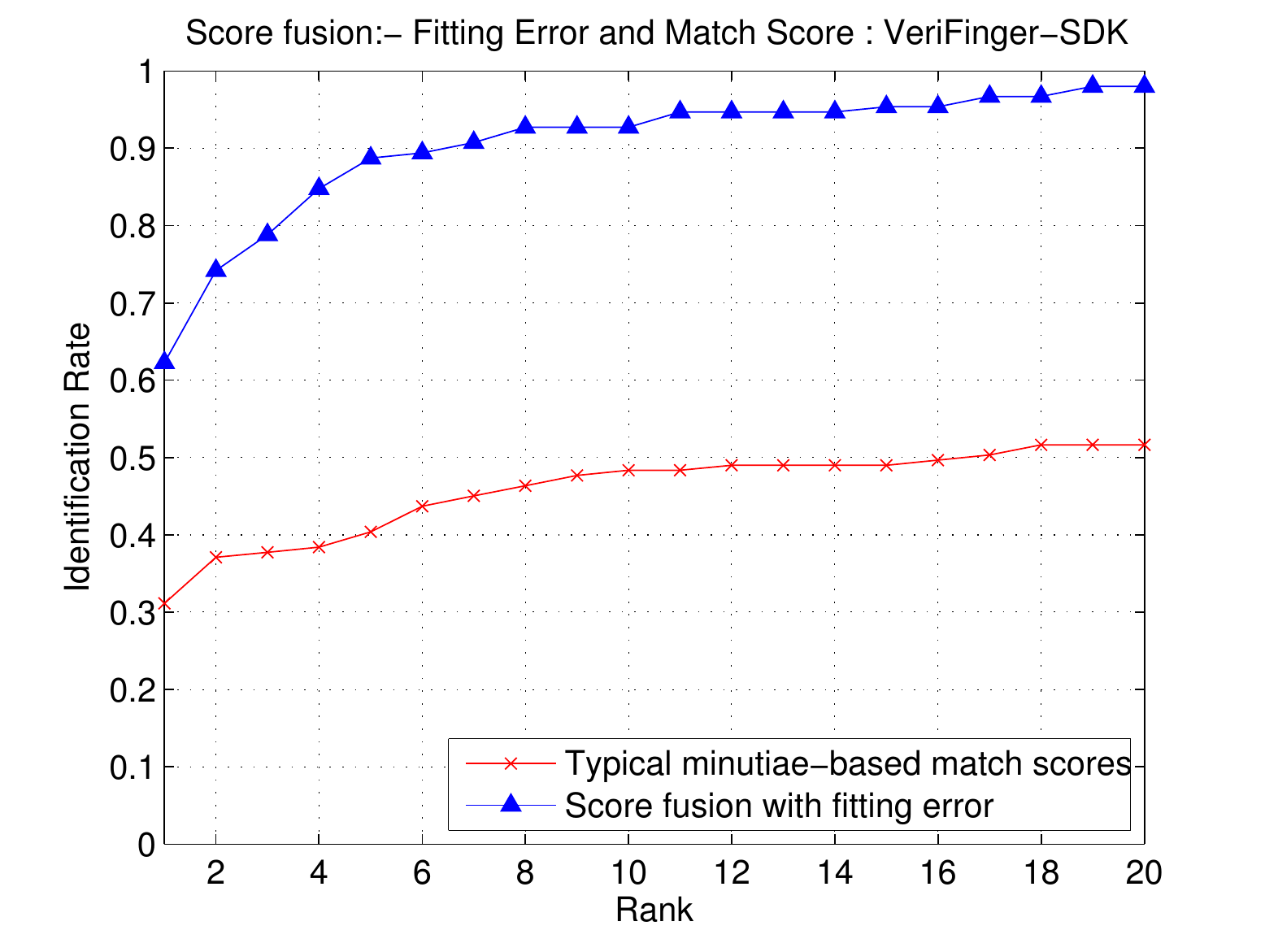}}
  \subfigure[CMC curve for MCC-SDK before and after score fusion]
            {\includegraphics[width=6cm]{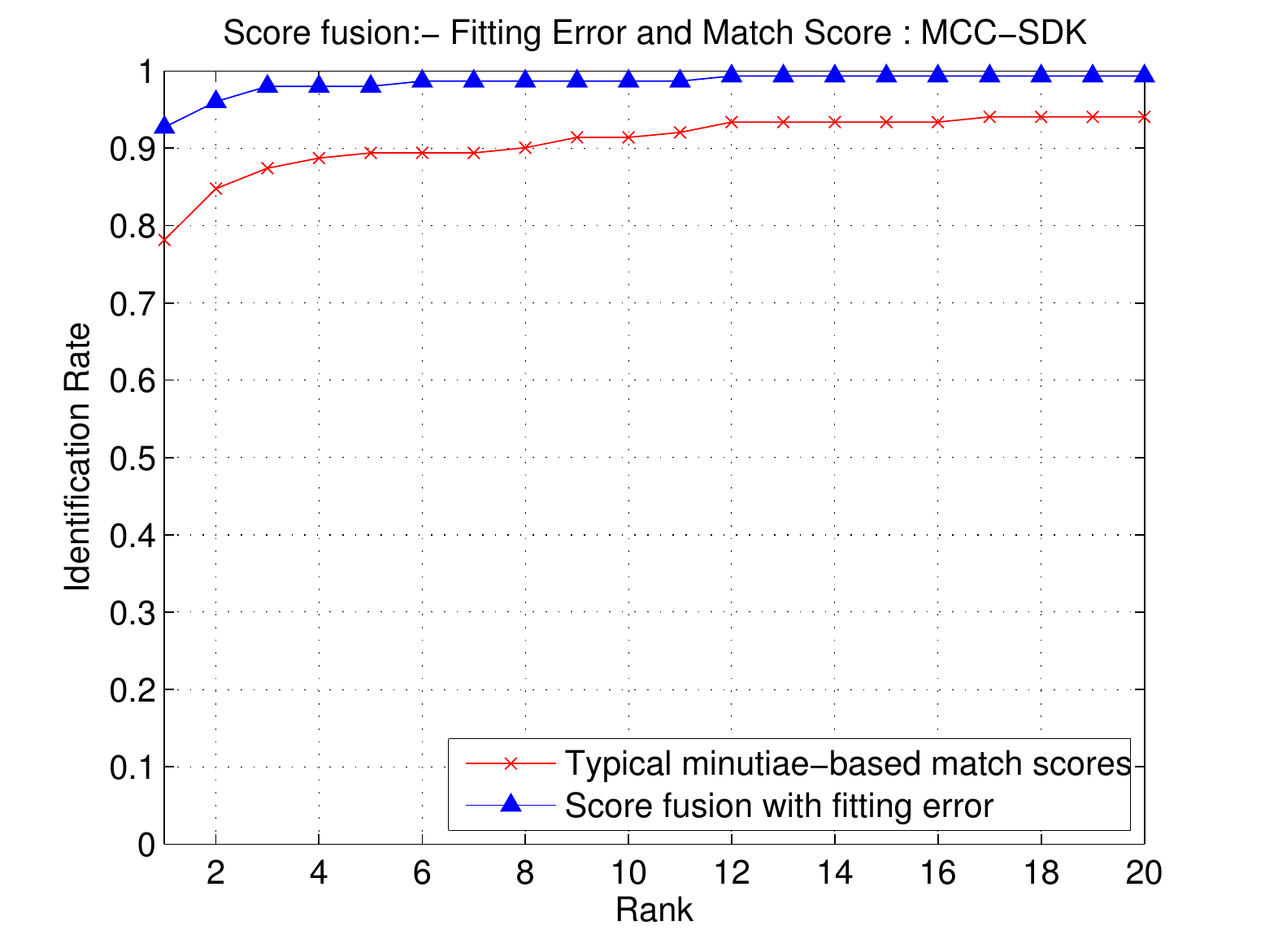}}
  \caption{CMC curve showing consistent improvement in the rank
    identification accuracies for NIST-Bozorth3, VeriFinger-SDK and
    MCC-SDK when their minutiae-based similarity scores
    are fused with fitting error based on the rare minutiae proposed
    in our algorithm.}
  \label{fig:exp2SF}
\end{center}
\end{figure}

Figures~\ref{fig:exp2SF}(a), ~\ref{fig:exp2SF}(b) and
~\ref{fig:exp2SF}(c) show the CMC curve
before and after score fusion for minutiae-based matchers
NIST-Bozorth3, VeriFinger-SDK and MCC-SDK respectively. Typical
minutiae-based similarity scores were very poor for NIST-Bozorth3 and
VeriFinger-SDK at Rank-1 identification and beyond. NIST-Bozorth3
achieved only $34.44\%$ Rank-1 identification accuracy when only
typical minutiae (ridge-endings and bifurcations) were
used. VeriFinger-SDK achieved $31.13\%$ Rank-1
identification accuracy under similar configuration.
The performance of MCC-SDK is far better than other two matchers, and
achieved $78.15\%$ Rank-1 identification accuracy when only using
typical minutiae. In general, MCC-SDK is one of the best performing
local structure based minutiae matcher.

\begin{table}[t]
  \footnotesize
\begin{center}
\begin{tabular}{|c|c|c|}
\hline
\bf{Matcher} & \bf{Before score fusion} & \bf{After score fusion} \\
& \bf{(Rank-1)} & \bf{(Rank-1)}\\

\hline\hline
NIST-Bozorth3 & 34.44 & 51.00\\
VeriFinger-SDK & 31.13  & 62.25\\
MCC-SDK & 78.15  & 92.72\\
\hline
\end{tabular}
\end{center}
\caption{Rank-1 identification (in \%) for NIST-Bozorth3,
  VeriFinger-SDK and MCC-SDK before and after score fusion.}
\label{tab:exp2}
\end{table}

The similarity score $S_m$ of the minutiae-based matcher is
modified by fusing with the fitting error
$\hat{E}^{\hat{L},\hat{M_s}}$ using mean rule to obtain $S'_m$
(Eq. (\ref{eq:fusion})). This score fusion significantly improves the
rank identification accuracies for all minutiae-based matchers. The
Rank-1 identification improved from $34.44\%$ to $51.00\%$ for
NIST-Bozorth3, $31.13\%$ to $62.25\%$ for VeriFinger-SDK, and $78.15\%$ to
$92.72\%$ for MCC-SDK. The improvement obtained at this stage is
significant and consistent at all the ranks
(Figures~\ref{fig:exp2SF}(a), ~\ref{fig:exp2SF}(b) and
~\ref{fig:exp2SF}(c)).

Table~\ref{tab:exp2} summarizes the Rank-1 accuracy for NIST-Bozorth3,
VeriFinger-SDK and MCC-SDK before and after score fusion with fitting
error obtained using our proposed method.

\subsection{Experiment 3: Score modification based on fitting error threshold}

The fitting error threshold $E_t$ plays a crucial factor in our proposed
algorithm. Based on the fitting error, a given comparison can be
concluded whether it is a match comparison or a non-match
comparison. If the fitting error $\hat{E}^{\hat{L},\hat{M_s}}$ is
greater than a threshold $E_t$, then the comparison is concluded a match
comparison. This is because $\hat{E}^{\hat{L},\hat{M_s}}$ is also
viewed as a similarity measure. When the comparison is deemed to be a
match comparison based on threshold $E_t$, then the fused score $S'_m$
is rewarded by multiplying with a constant $\alpha$. Similarly, when
the comparison is deemed as non-match, then fused score $S'_m$ is
penalized by multiplying with a constant $\beta$. In our experiments,
we empirically chose $\alpha = 2$ and
$\beta = 1$ (Step 10 of Algorithm, Eq.(\ref{eq:feThreshold})).

\begin{table}[t]
  \footnotesize
\begin{center}
\begin{tabular}{|c|c|c|c|}
\hline
\bf{Matcher} & \bf{Before score fusion} & \bf{After score fusion} & \bf{Threshold based}\\
& \bf{(Rank-1)} & \bf{(Rank-1)} & \bf{modification (Rank-1)}\\

\hline\hline
NIST-Bozorth3 & 34.44 & 51.00 & 74.83\\
VeriFinger-SDK & 31.13  & 62.25 & 77.48\\
MCC-SDK & 78.15  & 92.72 & 96.03\\
\hline
\end{tabular}
\end{center}
\caption{Rank-1 identification (in \%) for NIST-Bozorth3,
  VeriFinger-SDK and MCC-SDK when the fused scores are modified based
  on fitting error threshold. Results are reported on the subset of
  mated latent-tenprint pairs where rare features are found (151 out
  of 268 pairs).}
\label{tab:exp3}
\end{table}

\begin{figure*}
\begin{center}
  \subfigure[Looking for optimal threshold based on Rank-1
    identification accuracy for NIST-Bozorth3.]
            {\includegraphics[width=6cm]{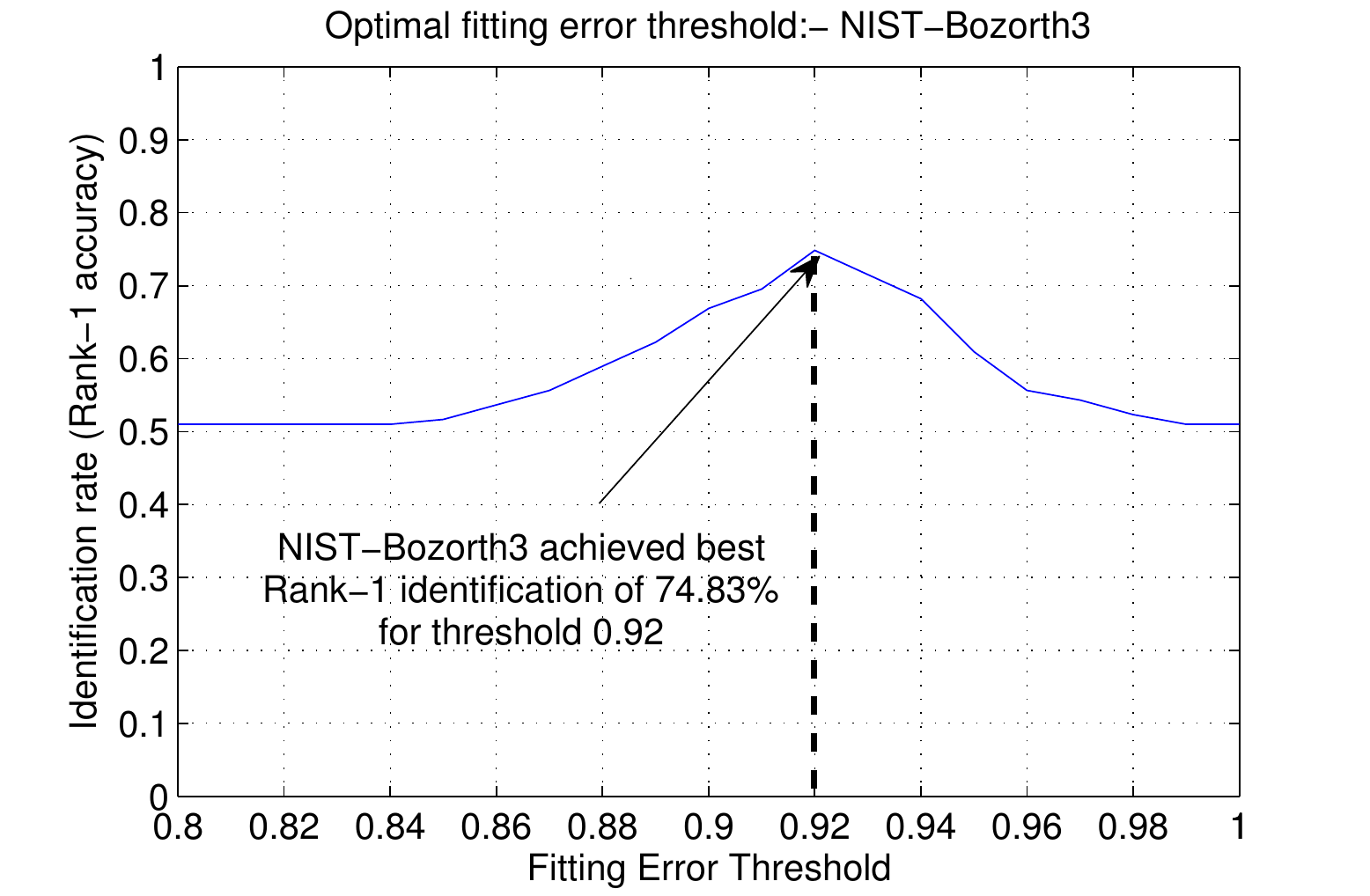}}
  \subfigure[CMC curve for NIST-Bozorth3 after modifying the fused
    score based on fitting error threshold.]
            {\includegraphics[width=6cm]{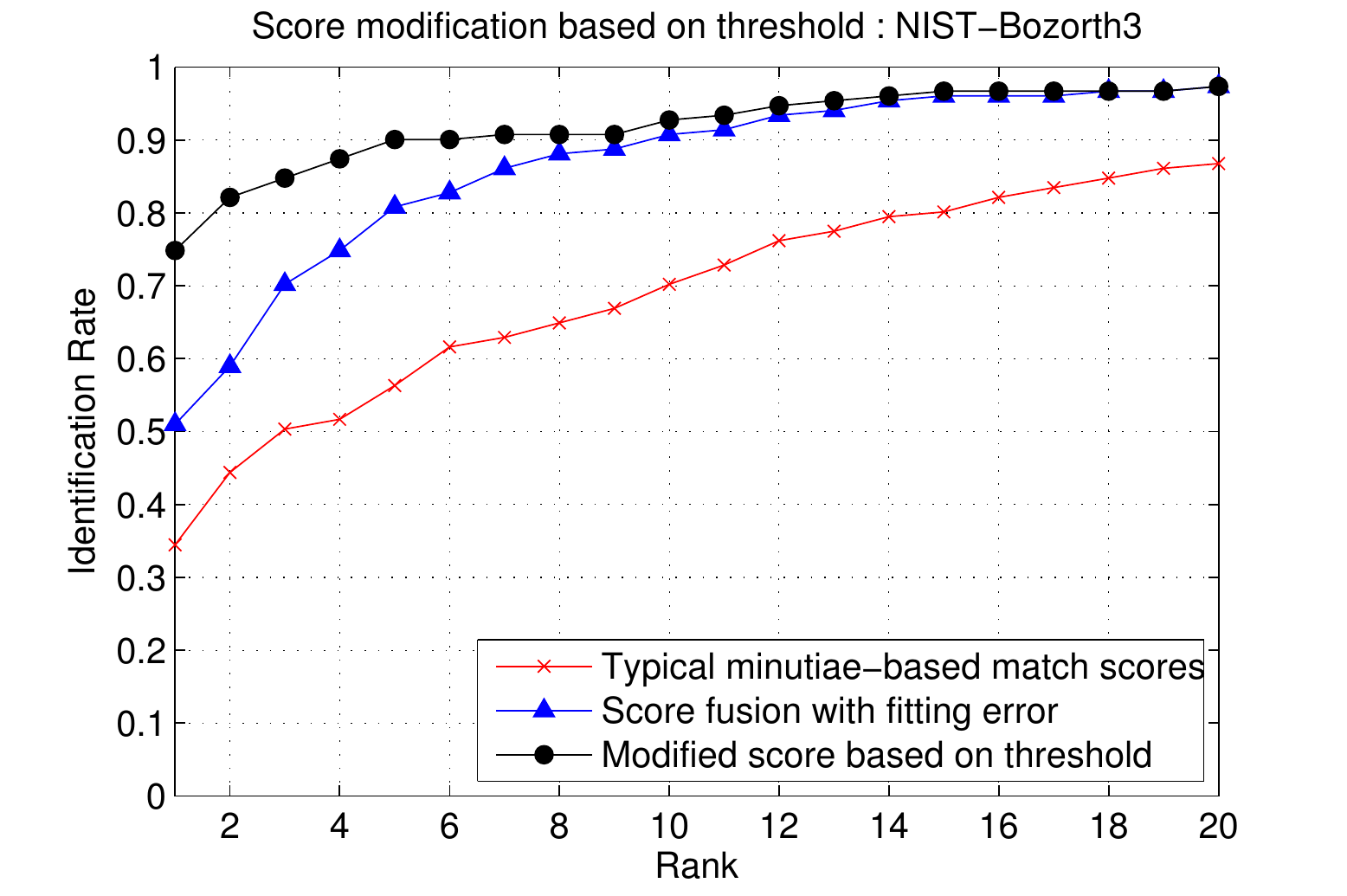}}
  \subfigure[Looking for optimal threshold based on Rank-1
    identification accuracy for VeriFinger-SDK.]
            {\includegraphics[width=6cm]{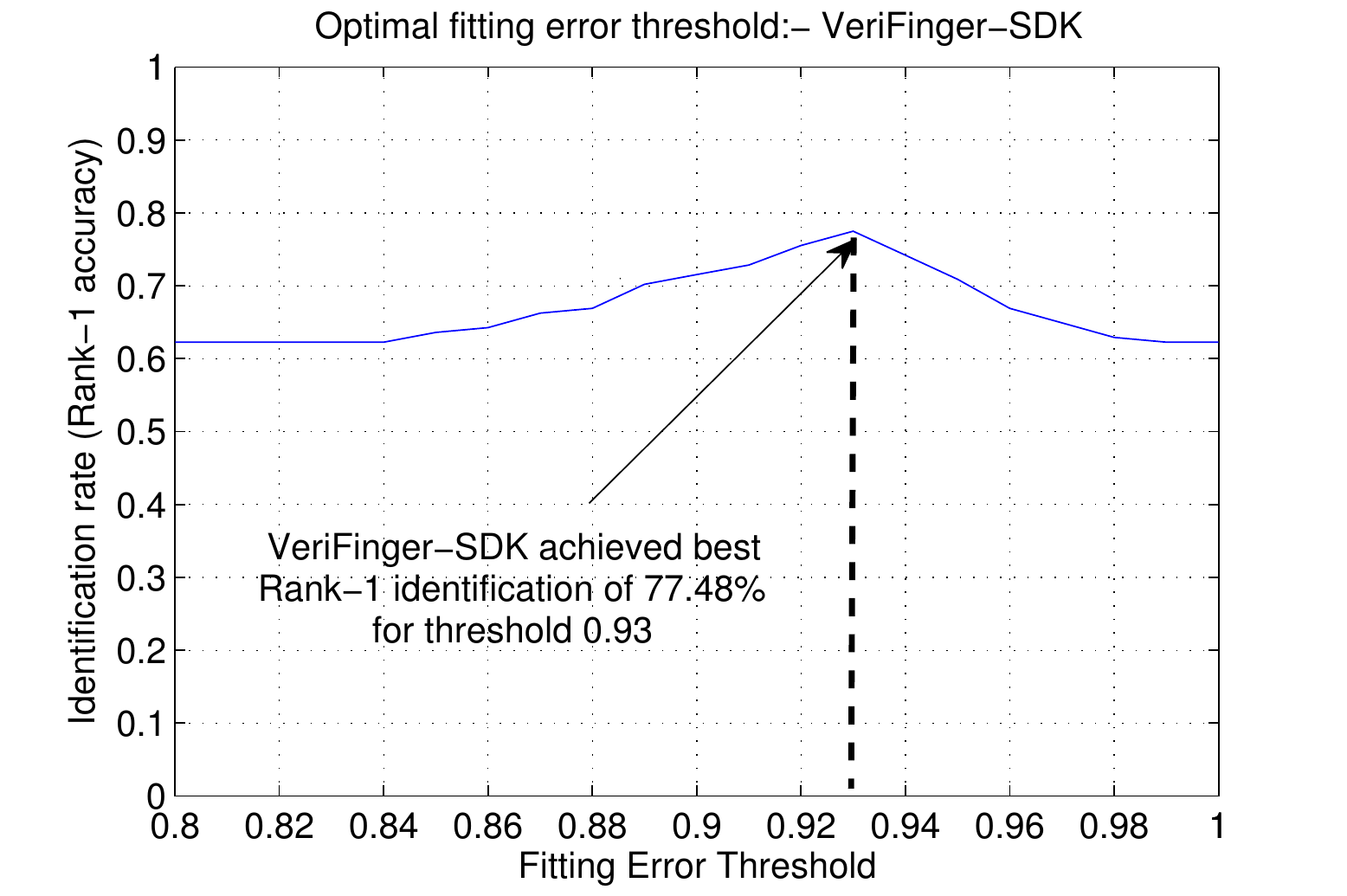}}
  \subfigure[CMC curve for VeriFinger-SDK after modifying the fused
    score based on fitting error threshold.]
            {\includegraphics[width=6cm]{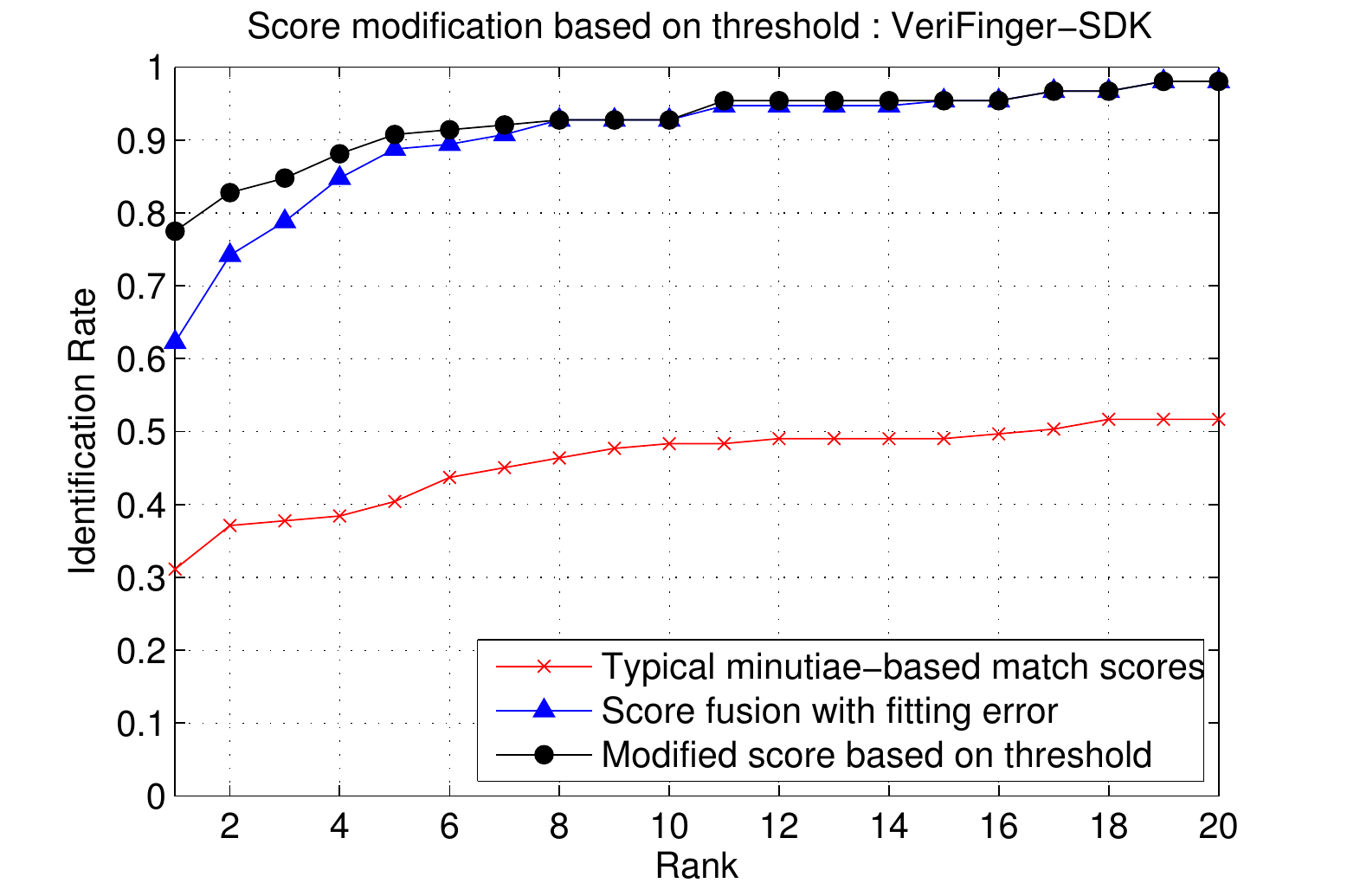}}
  \subfigure[Looking for optimal threshold based on Rank-1
    identification accuracy for MCC-SDK.]
            {\includegraphics[width=6cm]{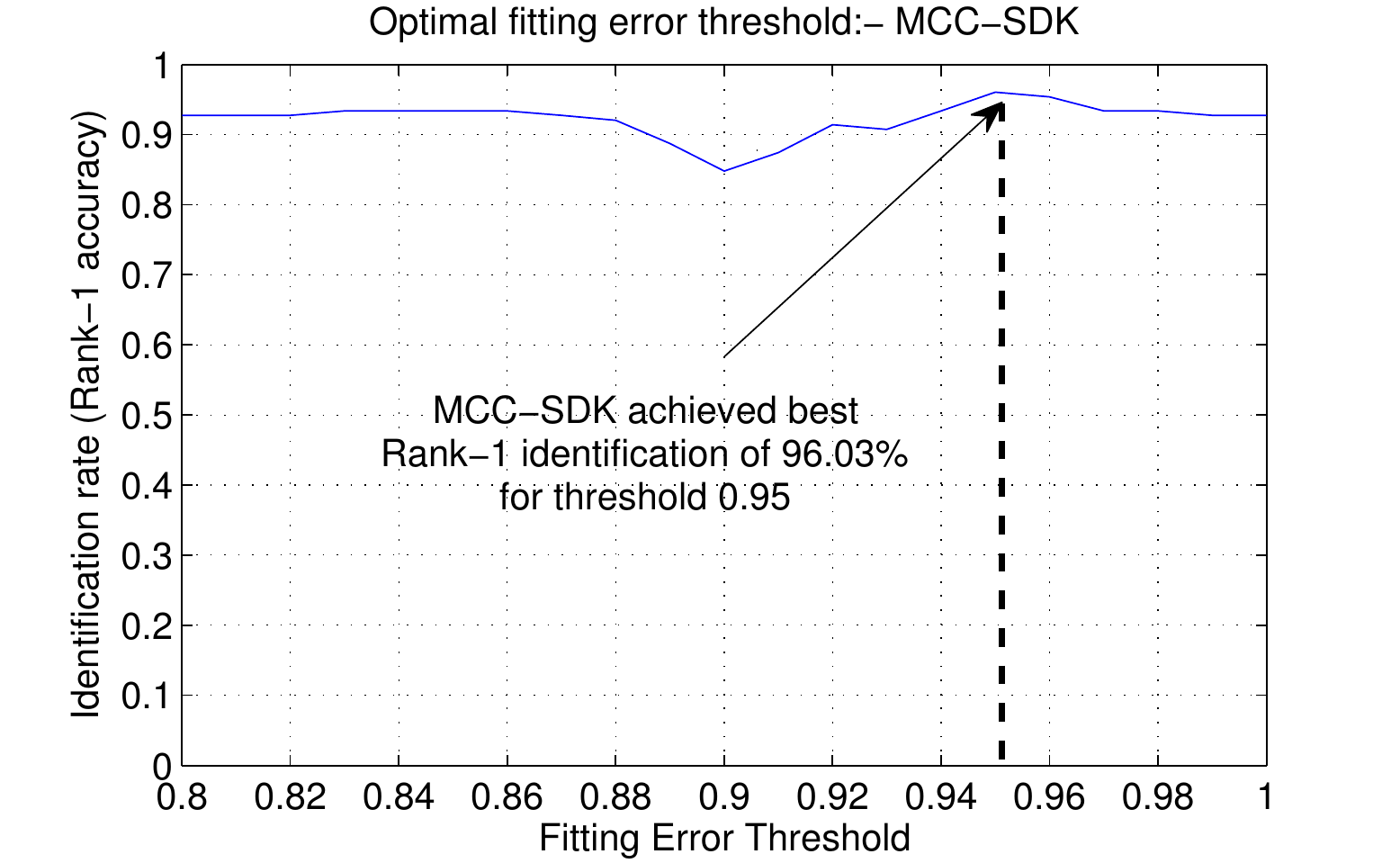}}
  \subfigure[CMC curve for MCC-SDK after modifying the fused
    score based on fitting error threshold.]
            {\includegraphics[width=6cm]{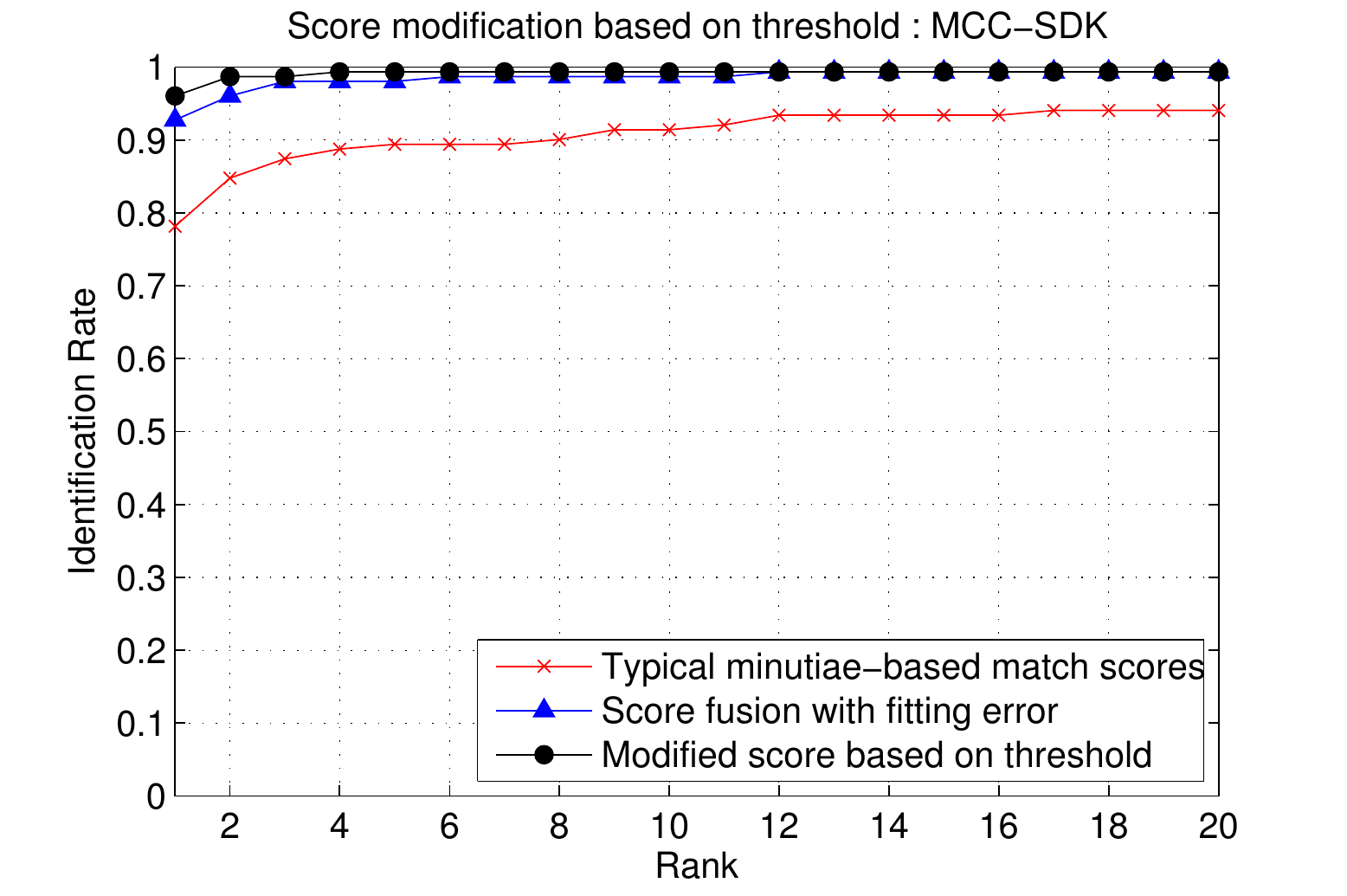}}
  \caption{CMC curve showing consistent improvement in the rank
    identification accuracies for NIST-Bozorth3, VeriFinger-SDK and
    MCC-SDK when the fused scores
    are further modified based on fitting error threshold. Left column
  show the parameter optimization by looking for optimal
  threshold. Right column show the improvement in rank identification
  after applying optimal threshold.}
  \label{fig:exp3B3}
\end{center}
\end{figure*}

Figures~\ref{fig:exp3B3}(a),~\ref{fig:exp3B3}(c) and~\ref{fig:exp3B3}(e)
show Rank-1 identification accuracies (Y-axis) obtained for various
fitting error thresholds (X-axis) for NIST-Bozorth3, VeriFinger-SDK
and MCC-SDK respectively.  The threshold is varied in the range $0.8$
to $1.0$, as most of the fitting error similarity values are
concentrated in this range (see Figure~\ref{fig:exp1}). The best
fitting error threshold for each of the matchers are obtained
heuristically. Same database is used for both this heuristic
estimation as well reporting the performance improvement of the
systems. This is particularly due to the objective of looking for best
possible performance the system can achieve as compared to only score
fusion. Consequently, this also
helps to understand the discriminating capability of the fitting
error, supporting Figure~\ref{fig:exp1}.

For NIST-Bozorth3, we notice that for a threshold value of $0.92$ (see
Figure~\ref{fig:exp3B3}(a)),
the system achieves $74.83\%$ Rank-1 identification accuracy. Using
our proposed algorithm, we were able to significantly improve the Rank-1
identification accuracy of NIST-Bozorth3 from $34.44\%$ to
$74.83\%$. Figure~\ref{fig:exp3B3}(b) show the CMC curve for
NIST-Bozorth3 when the fused scores are modified based on the fitting
error threshold. We observe a significant and consistent improvement
of rank identification accuracies for NIST-Bozorth3.

Figure~\ref{fig:exp3B3}(d) show the  CMC curve based on optimal
threshold for VeriFinger-SDK. For the optimal threshold
value of $0.93$ (see Figure~\ref{fig:exp3B3}(c)), VeriFinger-SDK
achieved Rank-1 identification
accuracy of $77.48\%$ which is a significant improvement from
$31.13\%$. We also observe that, using our proposed algorithm, we
achieve significant and consistent improvement in rank identification
accuracy for VeriFinger-SDK when the rare minutiae information is
incorporated.

Similarly, Figure~\ref{fig:exp3B3}(f)
show the best CMC curve for MCC-SDK. For an optimal
threshold value of $0.95$ (see Figure~\ref{fig:exp3B3}(e)), MCC-SDK
achieved the best Rank-1
identification of $96.03\%$, as well as consistent improvement in rank
identification using our proposed algorithm. Table~\ref{tab:exp3}
summarizes the Rank-1 identification for all the minutiae-based
matchers used in our experiments on the subset of mated
latent-tenprint pairs where rare features are found (151 out of 268
pairs).
Table~\ref{tab:expFULL} finally shows the performance over the full set of 268
latent-tenprint pairs, in which 151 out of them (the set that includes
rare features) are benefited from our proposed algorithm as indicated
in Table~\ref{tab:exp3}, and the remaining 117 are processed only with the baseline
AFIS under consideration (i.e., the similarity score $S_m$ in Figure~\ref{fig:algo}
is the only one used).

\begin{table}[t]
  \footnotesize
\begin{center}
\begin{tabular}{|c|c|c|}
\hline
\bf{Matcher} & \bf{Baseline AFIS} & \bf{Improved Matcher} \\
& \bf{(Rank-1)} & \bf{(Rank-1)}\\

\hline\hline
NIST-Bozorth3 & 25.37 & 36.94\\
VeriFinger-SDK & 31.72  & 59.33\\
MCC-SDK & 80.60  & 89.93\\
\hline
\end{tabular}
\end{center}
\caption{Rank-1 identification (in \%) for the 3 considered
  baseline fingerprint matchers before and after
applying the proposed improvement based on rare features, on the full
set of 268 latent-tenprint pairs included in GCDB database.}
\label{tab:expFULL}
\end{table}

%-------------------------------------------------------------------------

\section{Conclusions}\label{sect:conclusions}

We discussed the challenges faced by latent fingerprint
identification. One of the crucial challenges faced by AFIS is on
how to improve the rank identification accuracies when only partial
fingerprints are available.  We proposed an algorithm that
incorporates information from  reliably extracted rare minutia
features to improve the rank identification accuracies for minutiae
matchers.

The usefulness of the proposed algorithm is demonstrated on three widely
used minutiae-based matchers, NIST-Bozorth3, VeriFinger-SDK and MCC-SDK.
All the three  matchers showed significant
improvement in the rank identification accuracies when their
similarity scores were modified based on the
fitting error proposed in our methodology.  We conclude  that even if
we have only few number of minutiae in a
partial latent, presence of reliably extracted  rare minutia features
makes the comparison more robust.
In our experiments, we used the rare minutia features that were
manually extracted by forensic examiners.
Developing more robust automatic extraction of rare minutiae can
significantly improve the current state of the art in AFIS adapted for
latent fingerprints. Also, future work may exploit the differences
between minutiae types to further improve the performance of standard
AFIS, e.g., by incorporating weighting factors in the alignment or
matching steps~\cite{2018_INFFUS_MCSreview2_Fierrez} when combining
those AFIS with auxiliary approaches like the one presented in this paper.

%-------------------------------------------------------------------------

\section*{Acknowledgments}
R.K. was supported for the most part of this work by a Marie Curie
Fellowship under project BBfor2 from European Commission
(FP7-ITN-238803). This work has also been partially supported by
Spanish Guardia Civil, and project CogniMetrics (TEC2015-70627-R) from
Spanish MINECO/FEDER. The researchers from Halmstad University
acknowledge funding from KK-SIDUS-AIR project and the CAISR program in
Sweden.

%-------------------------------------------------------------------------
\section*{References}
{\small
  \bibliography{references}
}

\end{document}